\documentclass[lettersize,journal]{IEEEtran}
\usepackage{amsmath,amsfonts}
\usepackage{algorithmic}
\usepackage{algorithm}
\usepackage{array}
\usepackage[justification=centering]{caption}
\usepackage{subcaption}
\usepackage{textcomp}
\usepackage{multirow}
\usepackage{stfloats}
\usepackage{url}
\usepackage{xcolor}
\usepackage{verbatim}
\usepackage{graphicx}
\usepackage{cite}
\usepackage{amsthm}
\usepackage{textcomp}
\usepackage{stfloats}
\usepackage{url}
\usepackage{verbatim}
\usepackage{pifont}
\newtheorem{theorem}{Theorem}
\graphicspath{{Result/}}
\begin{document}

\title{SAR-FAH: A Frequency-Adaptive Hybrid Network based on Neural ODEs for Structural-Preserving SAR Despeckling}

\author{{Ziqing Ma},\ {Chang Yang},\ {Zhichang Guo},\ {Yao Li}\thanks{Corresponding author: Zhichang Guo (e-main: mathgzc@hit.edu.cn)}
\thanks{Ziqing Ma, Chang Yang, Zhichang Guo, and Yao Li are with the School of Mathematics, Harbin Institute of Technology, Harbin 150001, China. (e-mail: 23b912037@stu.hit.edu.cn, yangchang@hit.edu.cn, mathgzc@hit.edu.cn, yaoli0508@hit.edu.cn)}}


%

\maketitle

\begin{abstract}
Synthetic Aperture Radar (SAR) images are inherently degraded by speckle noise that severely limits their reliability in high-precision applications. As a signal-dependent multiplicative noise, speckle noise exhibits distinct statistical properties in homogeneous and heterogeneous regions of SAR images, which are spatially coupled. Nevertheless, existing deep learning despeckling methods operate directly in the spatial domain overlooking this statistical difference. It imposes a suboptimal trade-off between noise suppression and structure preservation, inevitably leading to artifacts, edge blurring, and texture distortion. To address these limitations, we propose a Frequency-Adaptive Hybrid model based on Neural Ordinary Differential Equations (NODEs) for SAR despeckling, termed SAR-FAH. It is a novel divide-and-conquer architecture that performs despeckling in the frequency domain to achieve improved structural preservation. We first fully decouple homogeneous and heterogeneous regions in the frequency domain via wavelet transform according to their local spatial characteristics and then revisit the statistical characteristics of speckle noise. Guided by the distinct properties of each sub-band, we design specialized sub-networks for frequency-specific restoration. Specifically, based on the smoothness of the low-frequency sub-band, the low-frequency denoising process is controlled by the module based on NODEs to ensure sufficient smoothness without artifacts, while the high-frequency sub-bands are processed by enhanced U-Net by incorporating deformable convolutions to better suppress noise and preserve edges and textures. Extensive experiments on both synthetic and real SAR images demonstrate that the proposed SAR-FAH outperforms the state-of-the-art methods both quantitatively and qualitatively.
\end{abstract}


\section{Introduction}
Synthetic Aperture Radar (SAR) is an all-weather, high-resolution coherent imaging sensor that captures terrain backscatter to produce images. Due to its unique imaging capabilities, it is widely used in military reconnaissance, terrain mapping, disaster monitoring, and various other applications \cite{frostModelRadarImages1982,cheneyMathematicalTutorialSynthetic2001}. However, the coherent interference of waves introduces speckle noise, a type of signal-dependent granular noise that typically manifests as bright or dark spots in the image. This strong coherence significantly disrupts image structures, making it difficult to perform downstream tasks such as segmentation \cite{luoMultiRegionSegmentationMethod2018}, recognition \cite{ohSpamnettarget2021}, and classification \cite{shiContentAdaptive2025}. Thus, speckle noise removal, also known as despeckling, while preserving important structures in SAR images, is a challenging yet significant task.\par 
Designing effective speckle reduction methods remains challenging due to the need to balance noise suppression with structural preservation. In recent years, deep learning approaches based on convolution neural networks (CNNs) have significantly advanced SAR despeckling by learning end-to-end mappings from noisy to clean images \cite{chierchiaSARImageDespeckling2017,zhangLearningDilatedResidual2018,koSARImageDespeckling2022}.  To better preserve structures, various frameworks have been developed, including residual learning \cite{zhangLearningDilatedResidual2018}, attention mechanisms \cite{koSARImageDespeckling2022}, gradient-based learning \cite{thakurAGSDNetAttentionGradientBased2022}, and transformer architectures \cite{pereraTransformerBasedSARImage2022}. Despite these advances, some limitations remain in effectively eliminating noise and preserving image structures.\par
\begin{figure}
\centering
\begin{minipage}{0.26\linewidth}
\begin{subfigure}{\linewidth}
\includegraphics[width=\textwidth]{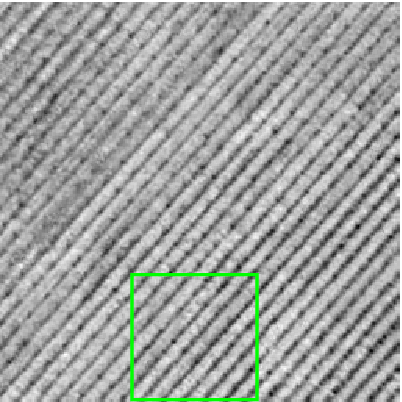}
\subcaption{}
\end{subfigure}
\end{minipage}
\begin{minipage}{0.12\linewidth}
\begin{subfigure}{\linewidth}
\includegraphics[width=\textwidth]{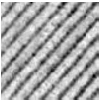}
\vspace{-0.26cm}
\\
\includegraphics[width=\textwidth]{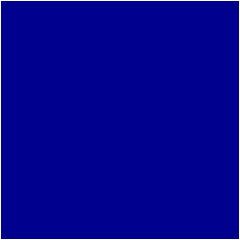}
\end{subfigure}
\subcaption{}
\end{minipage}
\begin{minipage}{0.12\linewidth}
\begin{subfigure}{\linewidth}
\includegraphics[width=\textwidth]{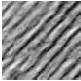}
\vspace{-0.26cm}
\\
\includegraphics[width=\textwidth]{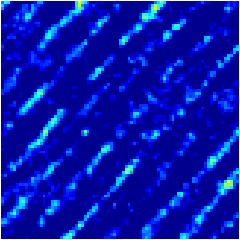}
\end{subfigure}
\subcaption{}
\end{minipage}
\begin{minipage}{0.12\linewidth}
\begin{subfigure}{\linewidth}
\includegraphics[width=\textwidth]{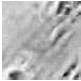}
\vspace{-0.26cm}
\\
\includegraphics[width=\textwidth]{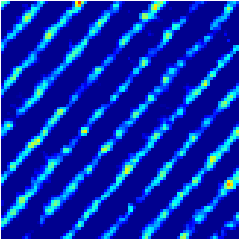}
\end{subfigure}
\subcaption{}
\end{minipage}
\begin{minipage}{0.12\linewidth}
\begin{subfigure}{\linewidth}
\includegraphics[width=\textwidth]{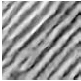}
\vspace{-0.26cm}
\\
\includegraphics[width=\textwidth]{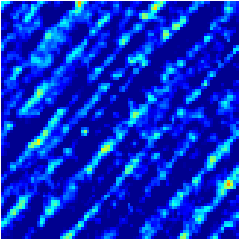}
\end{subfigure}
\subcaption{}
\end{minipage}
\begin{minipage}{0.12\linewidth}
\begin{subfigure}{\linewidth}
\includegraphics[width=\textwidth]{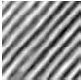}
\vspace{-0.26cm}
\\
\includegraphics[width=\textwidth]{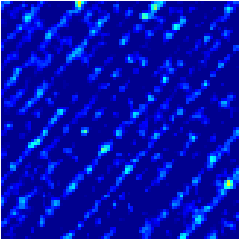}
\end{subfigure}
\subcaption{}
\label{fig:our}
\end{minipage}
\caption{An example of despeckling performance from the state-of-the-art methods to compare the restoration. (a) shows the clean SAR image. (b) shows the region in (a) boxed by green. (c - f) show despeckling results and their error maps of region in (a) boxed by green. (c) SAR-CAM \cite{koSARImageDespeckling2022}. (d) AGSDNet \cite{thakurAGSDNetAttentionGradientBased2022}. (e) SAR-TRN \cite{pereraTransformerBasedSARImage2022}. (f) SAR-FAH.}
\label{fig:dempose1}
\end{figure}

A key reason lies in the intrinsic noise statistical difference of SAR image. According to local spatial  characteristics, SAR images can be divided into homogeneous and heterogeneous regions. Homogeneous regions consist of pixels with slowly changing intensity, while heterogeneous regions contain areas with rich edges, textures, and isolated objects where local pixels change rapidly. Due to the signal-dependent nature of speckle, the noise follows distinct statistical distributions across these regions. Typically, speckle noise in homogeneous areas satisfies the fully developed assumption and follows a Gamma distribution \cite{freryModelExtremelyHeterogeneous1997}. Conversely, heterogeneous regions containing strong scatterers exhibit different noise characteristic and are better described by the $K$-distribution \cite{freryModelExtremelyHeterogeneous1997}. Moreover, homogeneous regions dominate the image energy. Since CNNs are inherently biased toward statistically dominant features  \cite{ayyoubzadehHighFrequencyDetail2021},  this imbalance causes them to prioritize learning homogeneous regions while inadequately capturing the distinct statistical properties and structures of heterogeneous areas. Consequently, existing models cannot effectively balance noise removal and structural preservation, especially for the denoising in heterogeneous regions, leading to edge blurring and textural distortion. As illustrated in figure \ref{fig:dempose1} with a farm SAR image, state-of-the-art methods exhibit fewer errors in homogeneous areas but fail to preserve edges and textures. The limited performance in heterogeneous regions hinders accurate detection of field boundaries and shapes.\par
To alleviate these problems, some studies have attempted to consider the statistical properties of speckle noise. For instance, a multi-objective loss function was designed to minimize the discrepancy between the ratio image and the Gamma-distributed noise, aiming to leverage the statistical properties of SAR images \cite{vitaleMultiObjectiveCNNBasedAlgorithm2021}. However, it ignores the distributional difference between homogeneous and heterogeneous regions, again leading to blurred edges and distorted textures. Furthermore, a structure extraction module was designed to extract heterogeneous features with weighting structural loss for collaborative optimization to protect edges and textural details \cite{yuCollaborativeOptimizationSAR2025a}. Nevertheless, because homogeneous and heterogeneous regions are spatially intertwined, such methods cannot fully decouple the structural information, resulting in a suboptimal recovery.\par
\begin{figure}[h]
	\centering
       \includegraphics[width=\linewidth]{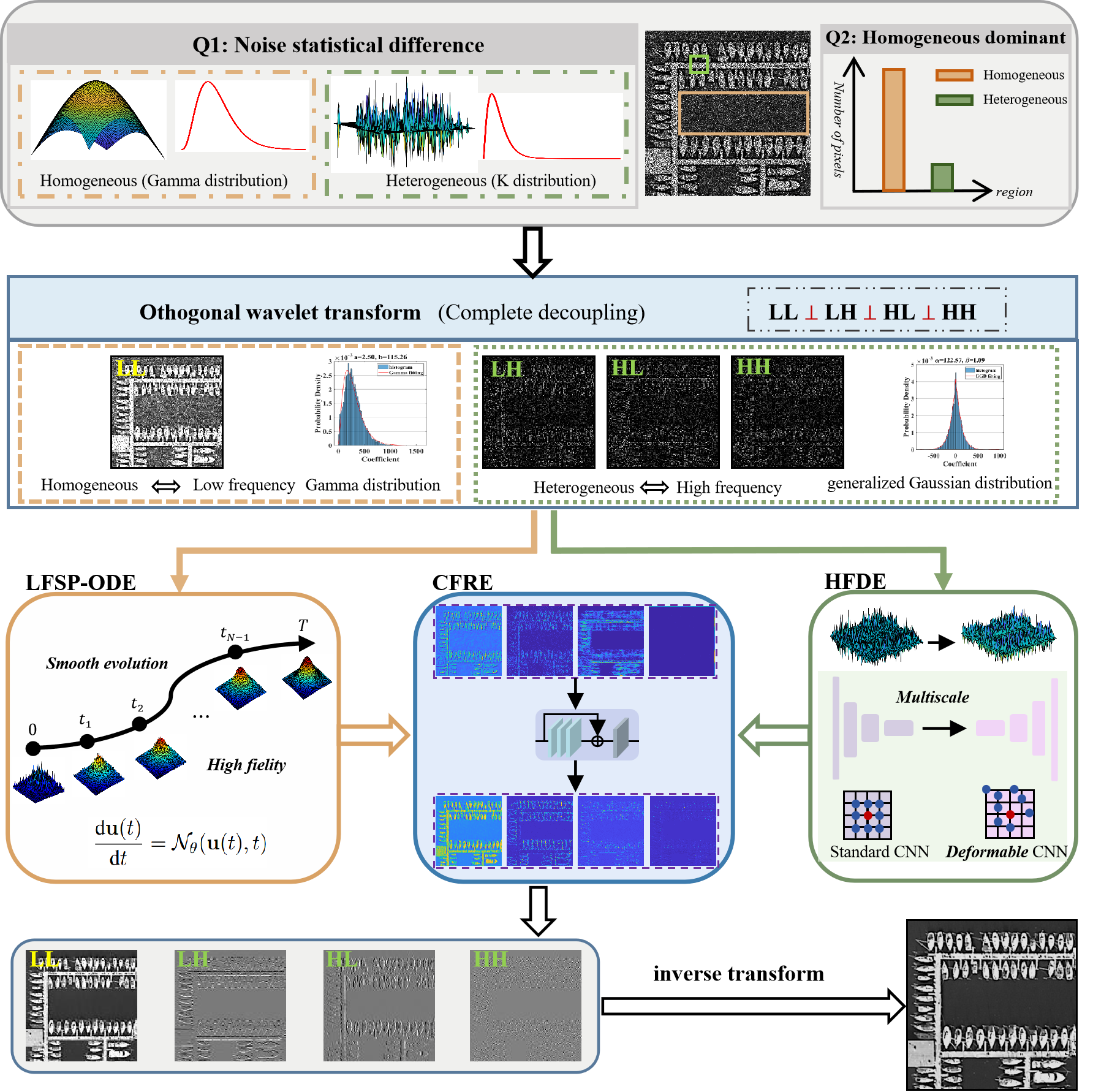}
	\caption{The overall framework of the proposed SAR-FAH model, which solves the problem caused by noise statistical difference and homogeneous dominant property. It include Low-Frequency Structural Preservation-ODE (LFSP-ODE) module, High-Frequency Denoising and Enhancement (HFDE) module, and Cross-FRequency Enhancement (CFRE) module.}\label{fig:process}
\end{figure}
In this paper, we propose SAR-FAH, which is a Frequency-Adaptive Hybrid model based on NODEs for SAR despeckling as illustrated in figure  \ref{fig:process}, aiming to balance noise suppression and structural preservation. Motivated by the local spatial characteristics of homogeneous and heterogeneous regions and the limitations caused by their spatial entanglement, we utilize the wavelet transform to decouple a SAR image into distinct frequency components. This explicit decomposition effectively separates regions dominated by different noise statistics and alleviates the issue of homogeneous region dominance. Also, it enables designing specialized sub-networks tailored to the denoising requirements of different sub-bands, thereby enhancing structural preservation. Specifically, for the low-frequency components that capture smooth homogeneous regions, we construct a Low-Frequency Structure-Preserving ODE (LFSP-ODE) module based on Neural Ordinary Differential Equations (NODEs) \cite{chenNeuralOrdinaryDifferential2018}. This module employs an ODE to govern the denoising process and leverages the feature extraction capability of CNNs to construct a continuous vector field, ensuring smooth evolution and structural fidelity without introducing artifacts. For the high-frequency sub-bands, which contain most of noise and sharp structures, we design a High-Frequency Denoising Enhancement (HFDE) module. It is built upon a U-Net \cite{ronnebergerUNetConvolutionalNetworks2015} architecture equipped with deformable convolutions (DeforConv) \cite{daiDeformableConvolutionalNetworks2017} to capture multi-scale features and adapt to irregular structures, thus effectively suppressing noise and eliminating edge blurring and textural distortion. Through this frequency-adaptive divide-and-conquer strategy, SAR-FAH achieves an excellent balance between noise removal and structural preservation, as verified in figure \ref{fig:our}.\par
To sum up, the main contributions of this work are as follows:
\begin{enumerate}
	\item We employ the wavelet transform to decouple SAR images into frequency sub-bands that completely separate homogeneous and heterogeneous regions. On this basis, we revisit the statistical differences of speckle noise between homogeneous and heterogeneous regions, and exploit these differences to enable frequency-adaptive despeckling.
	\item We propose SAR-FAH, a Frequency-Adaptive Hybrid network that follows a divide-and-conquer paradigm in the frequency domain. It assigns specialized sub-networks to low- and high-frequency sub-bands to effectively balance noise suppression and structural preservation.
	\item Leveraging the smoothness of low-frequency features, we design a LFSP-ODE module based on NODEs for artifact-free low-frequency restoration with enhanced structural fidelity. For high-frequency sub-bands, we construct a HFDE module with DeforConv that can effectively remove noise while preserving sharp high-frequency structures such as edges and textures. 
	\item Experiments on synthetic and real SAR datasets demonstrate that SAR-FAH achieves superior performance in SAR denoising. We further validate the excellent structure preservation capabilities of the proposed model in structure preservation through experiments on texture dataset.
\end{enumerate}
The remainder of this paper is structured as follows. Section \ref{sec:related} reviews related work on SAR despeckling. Section \ref{sec:proposed} illustrates our motivations and the proposed frequency-adaptive hybrid model in detail. Comprehensive experiments on validating despeckling performance are presented in Section \ref{sec:experiment}. Finally, we concludes in section \ref{sec:conclusion}.
\section{Related work}
\label{sec:related}
\subsection{Deep-learning SAR despeckling methods}
Advances in computing have driven the widespread adoption of deep learning in SAR despeckling. For instance, Chierchia et al. \cite{chierchiaSARImageDespeckling2017} combined logarithmic and exponential transformations to convert SAR despeckling into an additive noise removal task solving by DnCNN \cite{zhangGaussianDenoiserResidual2017}. Different this transform, Wang et al. \cite{wangSARImageDespeckling2017} employed similar network to estimate speckle noise, and combined total variation loss for end-to-end training enabling the preservation of edges. Furthermore, Zhang et al. \cite{zhangLearningDilatedResidual2018} applied dilated convolution and residual networks to enlarge the receptive field and mitigate the vanishing gradient problem in despeckling tasks.\par
Despite proficient noise suppression, existing CNN-based despeckling methods frequently exhibit suboptimal preservation of image structures including blurred edges and distorted textures. To bridge this gap, subsequent research has introduced more sophisticated architectures. Ko et al. \cite{koSARImageDespeckling2022} integrated an attention mechanism \cite{wooCBAMConvolutionalBlock2018} to refine features contextually, improving texture preservation. Liu et al. \cite{liuSpatialTransformDomain2022} decomposed the image via wavelet transform, using attention to suppress noise in frequency sub-bands, complemented by multi-scale feature extraction for structural enhancement. Additionally, Li et al. \cite{liuMFAENetMultiscaleFeature2023} embedded multi-scale extraction within a U-shape framework for structural similar enhancement. To explicitly guide the network in preserving edges, Thakur et al. \cite{thakurAGSDNetAttentionGradientBased2022} augmented features with gradient information to sharpen edges. Acknowledging the fundamental limitation of the local receptive field of CNNs, Perera et al. \cite{pereraTransformerBasedSARImage2022} introduced a transformer-based model, leveraging self-attention to capture long-range dependencies and generating high-resolution details as well as low-resolution coarse structures.\par
Nevertheless, the aforementioned approaches overlook the inherent statistical properties in homogeneous and heterogeneous regions, often leading to artifacts or the loss of sharp structures. In response, Vitale et al. \cite{vitaleMultiObjectiveCNNBasedAlgorithm2021} formulated a multi-objective loss function aimed at preserving statistical properties on homogeneous regions, edges, and textures simultaneously. More recently, Yu et al. \cite{yuCollaborativeOptimizationSAR2025a} utilized a pre-trained edge detection network to extract heterogeneous features, providing structural supervision by collaborative optimization. Despite these efforts, they still failed to effectively exploit the spatial statistical difference of SAR images due to the entanglement of these regions in spatial domain, resulting in suboptimal despeckling results.\par
\subsection{NODEs for structural fidelity}
While residual networks have enabled the construction of deeper architectures for stronger feature extraction, their inherently discrete block-by-block processing often leads to structural degradation and a rapid explosion in parameter count, which fundamentally limits their effectiveness in SAR despeckling tasks. To overcome these limitations, NODE \cite{chenNeuralOrdinaryDifferential2018} presents a compelling alternative by modelling feature evolution through a continuous dynamical system, which has shown promising results in related image restoration tasks \cite{xieNODEImgNetPDEinformedEffective2024b,yangSpikeODEreconstruction2024}. It is expressed as follows:
\begin{equation}
\medmuskip=0mu
\thickmuskip=0mu
\thinmuskip=0mu
	\begin{aligned}
		\frac{\text{d}\mathbf{u}(t)}{\text{d}t}  & = \mathcal{N}_{\theta}(\mathbf{u}(t), t),\ t\in\left(0,T\right],\\
	                               \mathbf{u}(0) & = \mathbf{f},
	\end{aligned}
	\label{eq:node}
\end{equation}
where the function $\mathcal{N}_{\theta}$ parameterizes the time derivative of the hidden state $\mathbf{u}(t)$ using a neural network, learning the continuous dynamics that govern the feature evolution. It inherits advantageous properties of ODE solutions, ensuring structural fidelity. In practice, the system is discretized and solved numerically using an ODE solver. As the complexity of the network increases, the parameter count remains unchanged. On the contrary, with the number of times solved by the ODE solver, the accuracy of the solution continues to improve with increased computational complexity obtaining better parameter utilization.
\subsection{VMamba for global feature extraction}
Global feature extraction plays a crucial role in preserving textural structures in images. While transformer-based models excel at capturing global contextual information, their computational complexity increases quadratically with sequence length, hindering practical scalability. To overcome this limitation, Liu et al. \cite{liuVMambaVisualState2024} introduced vision mamba (VMamba), a model built upon the Mamba framework \cite{guMambaLinearTimeSequence}, which enables efficient long-range dependency modelling with linear computational complexity. Its efficiency has proven beneficial across various image processing tasks \cite{shiVmambaIRVisualState2025,ZhouMaDiNetdiffusiondetection2025,duMLMambaclassification2025}. VMamba incorporates a 2-dimensional (2D) selective scanning (SS2D) mechanism, which converts image data into 1D sequences. These sequences are then processed by a selective state space model (SSM) \cite{guEfficientlyModelingLong2022a} to capture global dependencies. The model subsequently applies a weighting mechanism to reconstruct the 2D structural information from the outputs. The SSM maps a 1D input $x(s)\in \mathbb{R}$ to an output $y(s)\in \mathbb{R}$ via a hidden state $\mathbf{h}(s)\in \mathbb{R}^{d}$, where $L$ denotes the hidden state dimension, respectively. The continuous-time SSM is described by a linear ODE system:
\begin{equation*}
\medmuskip=0mu
\thickmuskip=0mu
\thinmuskip=0mu
	\mathbf{h}'(s)=\mathbf{A}\mathbf{h}(s)+\mathbf{B}x(s),\ y(s)=\mathbf{C}\mathbf{h}(s)+Dx(s),
\end{equation*} 
where $\mathbf{A}\in \mathbb{R}^{d\times d}$ and $D\in \mathbb{R}$ are learnable parameters, while $\mathbf{B}\in \mathbb{R}^{d\times 1}$ and $\mathbf{C}\in \mathbb{R}^{1\times d}$ are derived from the input $x(s)$. To integrate this system into deep learning frameworks, the continuous-time model is discretized using a learnable timescale parameter $\mathbf{\Delta} \in \mathbb{R}^{d\times d}$. Applying the Zero-Order Hold method \cite{pechlivanidouZeroorderHoldDiscretization2022} yields the discrete form:
\begin{equation}
\medmuskip=0mu
\thickmuskip=0mu
\thinmuskip=0mu
\begin{aligned}
	\bar{\mathbf{A}}=&e^{\mathbf{\Delta} \mathbf{A}},\ 
	\bar{\mathbf{B}}=(\mathbf{\Delta} \mathbf{A})^{-1}(e^{\mathbf{\Delta}\mathbf{A}}-\mathbf{I})\cdot \mathbf{\Delta} \mathbf{B},\\
	\mathbf{h}_s&=\bar{\mathbf{A}}\mathbf{h}_{s-1}+\bar{\mathbf{B}}x_s,\ y_s=\mathbf{C}\mathbf{h}_s+Dx_s.
	\end{aligned}
	\label{eq:ssm}
\end{equation}\par
\section{Proposed method}
\label{sec:proposed}
\subsection{Motivations}
\begin{figure}[hbtp]
\centering
\begin{subfigure}{0.25\linewidth}
\centering
\includegraphics[width=\textwidth]{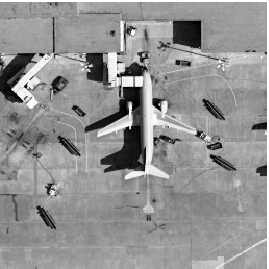}
\subcaption{\scriptsize SAR image}
\end{subfigure}
\begin{subfigure}{0.26\linewidth}
\centering
\includegraphics[width=\textwidth]{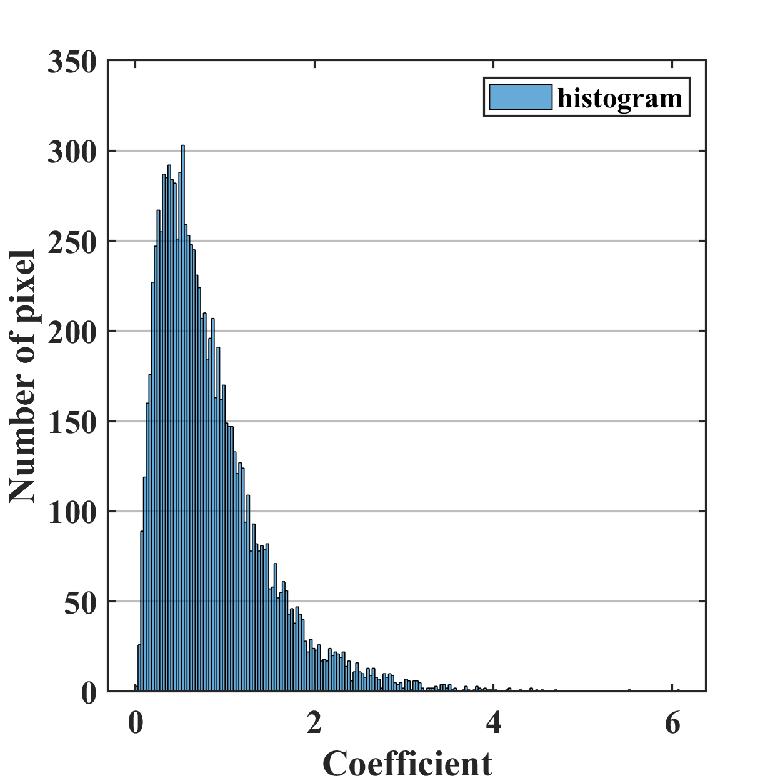}
\subcaption{\scriptsize homogeneous}
\end{subfigure}
\begin{subfigure}{0.26\linewidth}
\centering
\includegraphics[width=\textwidth]{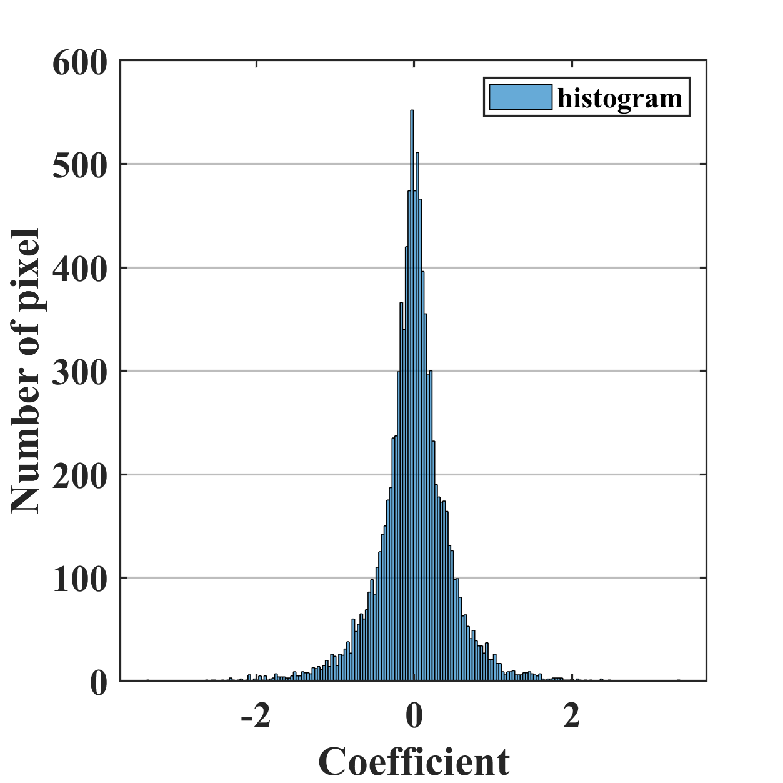}
\subcaption{\scriptsize heterogeneous}
\end{subfigure}
\caption{Energy analysis of SAR image. It illustrates the homogeneous region dominance property in SAR image.}
\label{fig:databias}
\end{figure}

SAR despeckling aims to balance noise removal and structural preservation. Existing deep-learning methods often struggle to preserve the structures of SAR images during noise smoothing, resulting in blurred edges and distorted textures. Deep learning models, as data-fitting tools that minimize a global loss function via gradient descent, inherently tend to fit the majority of samples during training to rapidly reduce overall loss \cite{ayyoubzadehHighFrequencyDetail2021}. SAR images can be categorized into homogeneous and heterogeneous regions based on local spatial characteristics. Usually, there is a significant difference in the proportion of these two parts. In detail, homogeneous areas dominate the image, while heterogeneous structures such as edges and textures occupy only a small fraction. As illustrated in figure \ref{fig:databias}, an analysis of the energy distribution reveals a clear energy bias that heterogeneous regions contribute far less total energy than homogeneous ones as its energy is concentrated at $0$. Consequently, deep models naturally achieve better denoising performance in homogeneous regions but tend to underfit heterogeneous structures, leading to blurred edges and distorted textures. Therefore, explicitly addressing this energy bias is essential for preserving sharp structures.\par
Spatial imbalance further limits the despeckling performance of deep learning models due to the difference of the speckle statistical distribution. Speckle noise in SAR images is typically modelled as multiplicative noise. Under the well-developed hypothesis, it follows a Gamma distribution, denoted by $\mathcal{GM}$, characterized by the probability density function: 
\begin{equation*}
\medmuskip=0mu
\thickmuskip=0mu
\thinmuskip=0mu
	p(x;a,b)=\frac{b^{-a}}{\Gamma(a)}x^{a-1} \exp(-\frac{x}{b})\chi_{\{x \geq 0\}},
\end{equation*}
where $\chi$ is the indicator function, $a$ and $b$ express shape and scale parameter, respectively. Usually, it is supposed that mean $ab$ is equal to 1, and the variance if $ab^2=L$, where $L$ is number of looks.\par
However, in heterogeneous regions containing strong scatterers, the conditions for fully developed speckle are often not satisfied. As a result, these regions are better characterized by a $K$-distribution \cite{freryModelExtremelyHeterogeneous1997}. This divergence in statistical characteristics between homogeneous and heterogeneous areas poses a significant challenge for despeckling models. A single model struggles to accommodate both statistical regimes effectively, leading to a problematic ''compromise'' in a single complex mapping task. Typically, it faces a dilemma. If it focuses on fully suppressing noise, excessive smoothing tends to blur edges and distort textures. Conversely, if it prioritizes preserving structural features in heterogeneous regions, its limited smoothing ability often leaves substantial residual noise and artifacts in uniform areas. The fundamental issue is that these two regions are often spatially intertwined, preventing existing models from fully leveraging their inherent statistical differences and thereby limiting overall despeckling performance.\par
A key observation is that homogeneous regions exhibit gentle intensity transitions, while heterogeneous regions correspond to rapid transitions. Consequently, they possess inherent frequency differences, which drives us to effectively separate them in the frequency domain, thereby jointly addressing the energy bias and exploiting the distinct noise statistics. Specifically, in this paper, we address this by integrating wavelet decomposition, particularly the 2D separable Haar wavelet transform (2D-HWT). Grounded in multi-resolution analysis theory \cite{mallatTheoryMultiresolutionSignal1989} for $L^2(\mathbb{R}^2)$, this approach projects an image onto different orthogonal subspaces. It effectively separates features and elevates them to equal importance, thereby solving the problem of imbalanced proportion. Specifically, the image is decoupled into four physically distinct subspaces, including a low-frequency subspace that captures information from homogeneous regions, and three high-frequency subspaces that focus on heterogeneous, directionally-oriented features. This decomposition creates a more balanced feature learning process, encouraging the model to perform comprehensive restoration across all image regions.\par
The 2D separable orthonormal Haar wavelet filters are defined as:
\begin{equation}
\medmuskip=0mu
\thickmuskip=0mu
\thinmuskip=0mu
\begin{aligned}
\boldsymbol{\omega}^{\text{LL}} &=\frac{1}{2}
\begin{bmatrix} 1 & 1 \\ 1 & 1 
\end{bmatrix},\ 
\boldsymbol{\omega}^{\text{LH}}=\frac{1}{2}
\begin{bmatrix} -1 & -1 \\ 1 & 1 
\end{bmatrix},\\
\boldsymbol{\omega}^{\text{HL}}&=\frac{1}{2}
\begin{bmatrix} -1 & 1 \\ -1 & 1 
\end{bmatrix},\ 
\boldsymbol{\omega}^{\text{HH}}=\frac{1}{2}
\begin{bmatrix} 1 & -1 \\ -1 & 1 
\end{bmatrix}.
\end{aligned}
\label{eq:wavelet}
\end{equation} 
With these filters, homogeneous regions are projected into the low-frequency sub-band via $\boldsymbol{\omega}^{\text{LL}}$, while heterogeneous structures are mapped into high-frequency sub-bands in three directions by $\{\boldsymbol{\omega}^{\text{LH}},\boldsymbol{\omega}^{\text{HL}},\boldsymbol{\omega}^{\text{HH}}\}$. After this decomposition, the spatial differences in the noise of SAR images are transformed into frequency domain differences. And through the theorem \ref{the:low gamma} referred to \cite{cohenDesignExperimentsStatistical2016}, we thoroughly analyze the statistical distributions of the coefficients in different sub-bands.\par
\begin{figure*}
\centering
\begin{subfigure}{0.24\linewidth}
\centering
\includegraphics[width=0.45\textwidth]{Contri/gt.eps}
\includegraphics[width=0.45\textwidth]{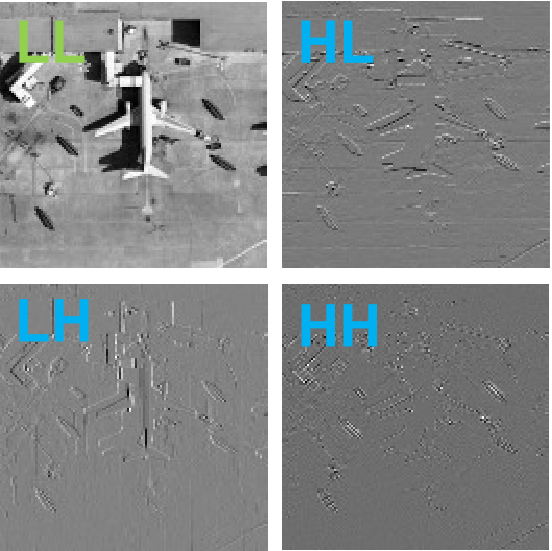}
\subcaption{\scriptsize Clean}
\end{subfigure}
\hspace{0.5cm}
\begin{subfigure}{0.24\linewidth}
\centering
\includegraphics[width=0.45\textwidth]{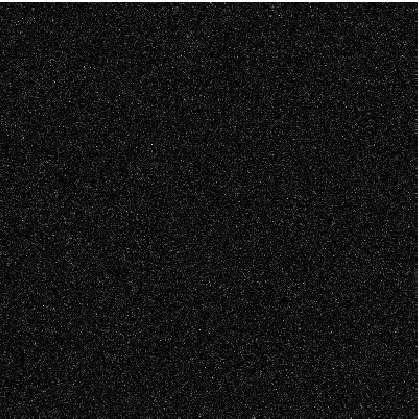}
\includegraphics[width=0.48\textwidth]{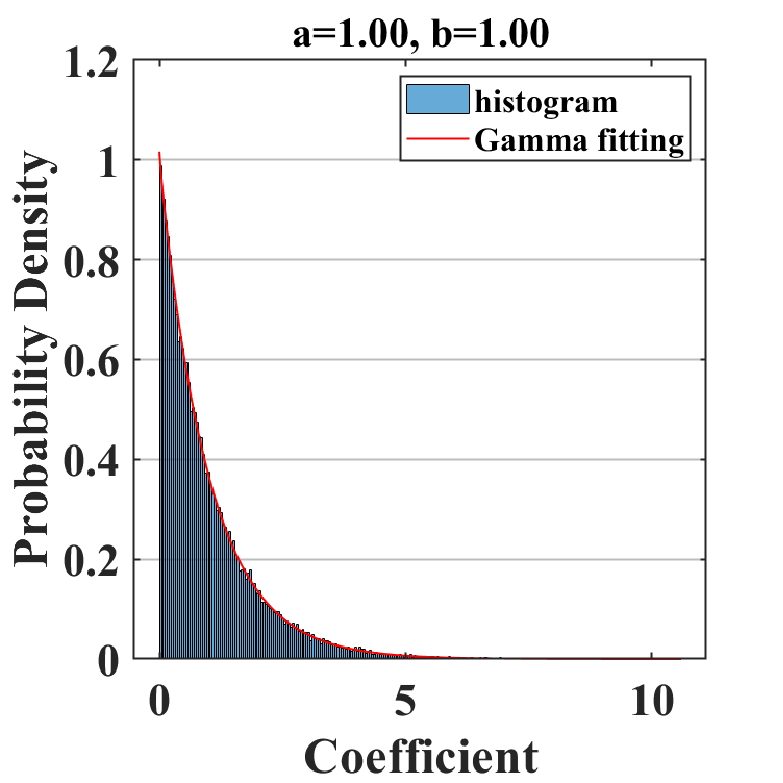}
\subcaption{\scriptsize Speckle noise ($L=1$)}
\end{subfigure}
\hspace{0.5cm}
\begin{subfigure}{0.24\linewidth}
\centering
\includegraphics[width=0.45\textwidth]{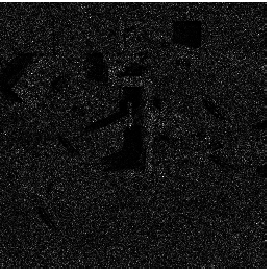}
\includegraphics[width=0.48\textwidth]{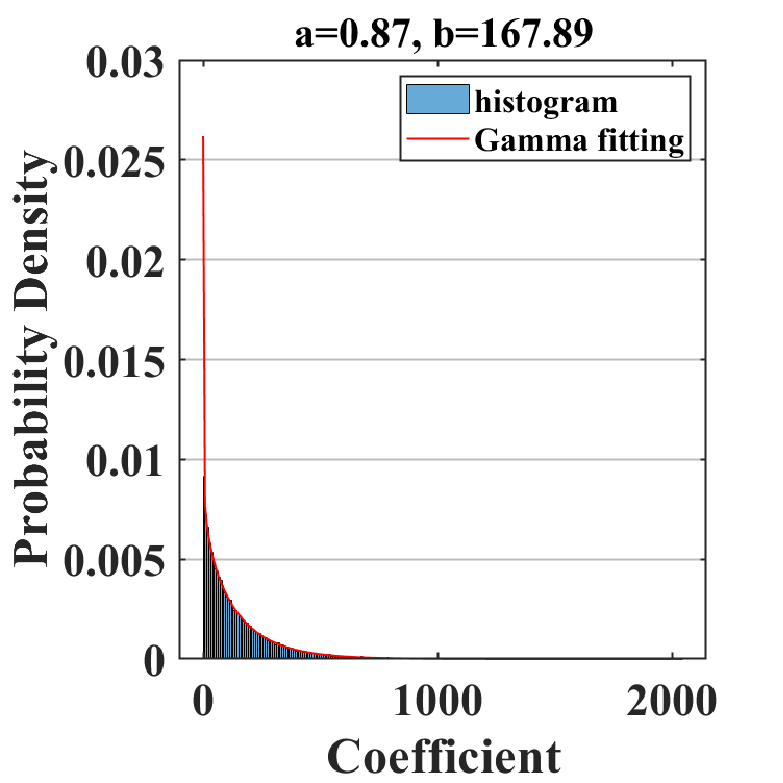}
\subcaption{\scriptsize Noisy (PSNR=8.89, SSIM=0.0671)}
\end{subfigure}
\\
\begin{subfigure}{0.24\linewidth}
\includegraphics[width=0.45\textwidth]{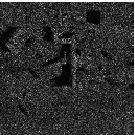}
\hspace{0.05cm}
\includegraphics[width=0.48\textwidth]{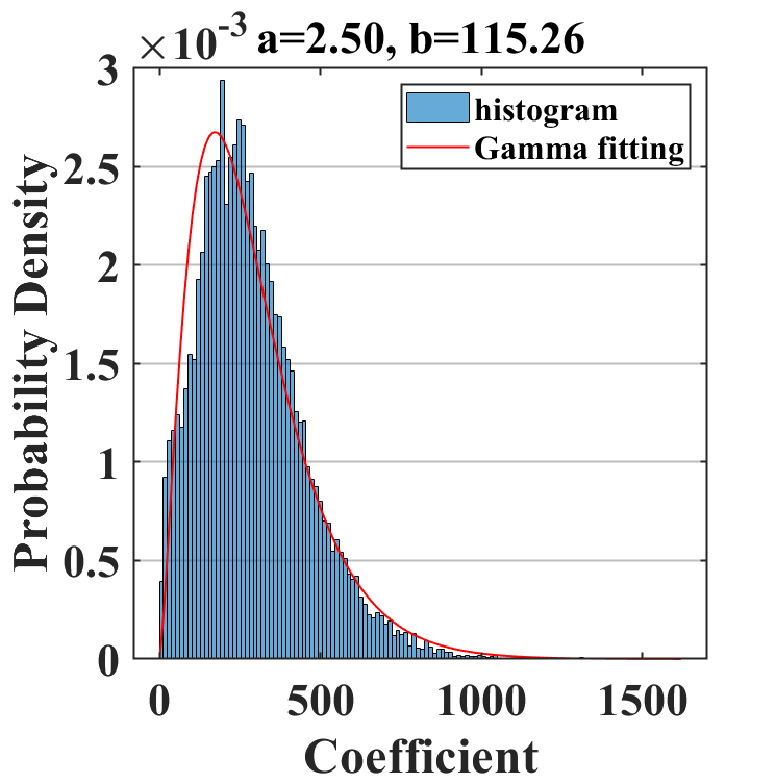}
\subcaption{\scriptsize $\mathbf{X}^{\text{nLL}}$ (PSNR=12.34, SSIM=0.2255)}
\end{subfigure} 
\begin{subfigure}{0.24\linewidth}
\centering
  \includegraphics[width=0.45\textwidth]{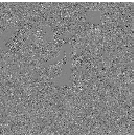}
  \hspace{0.05cm}
  \includegraphics[width=0.48\textwidth]{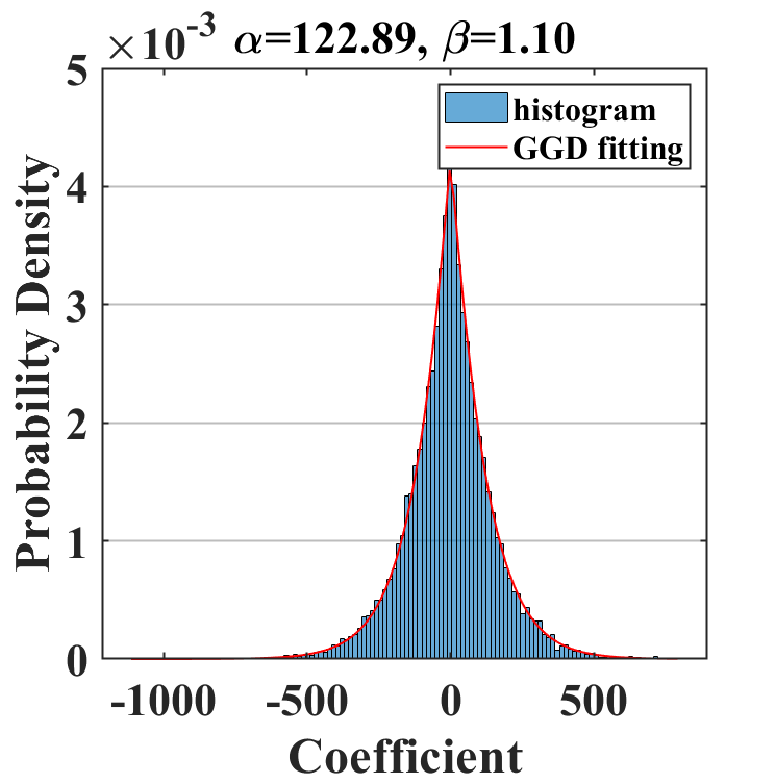}
    \subcaption{\scriptsize $\mathbf{X}^{\text{nLH}}$ (PSNR=6.56, SSIM=0.0177)}
    \end{subfigure}
\begin{subfigure}{0.24\linewidth}
\centering
  \includegraphics[width=0.45\textwidth]{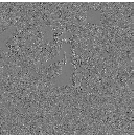}
  \hspace{0.05cm}
  \includegraphics[width=0.48\textwidth]{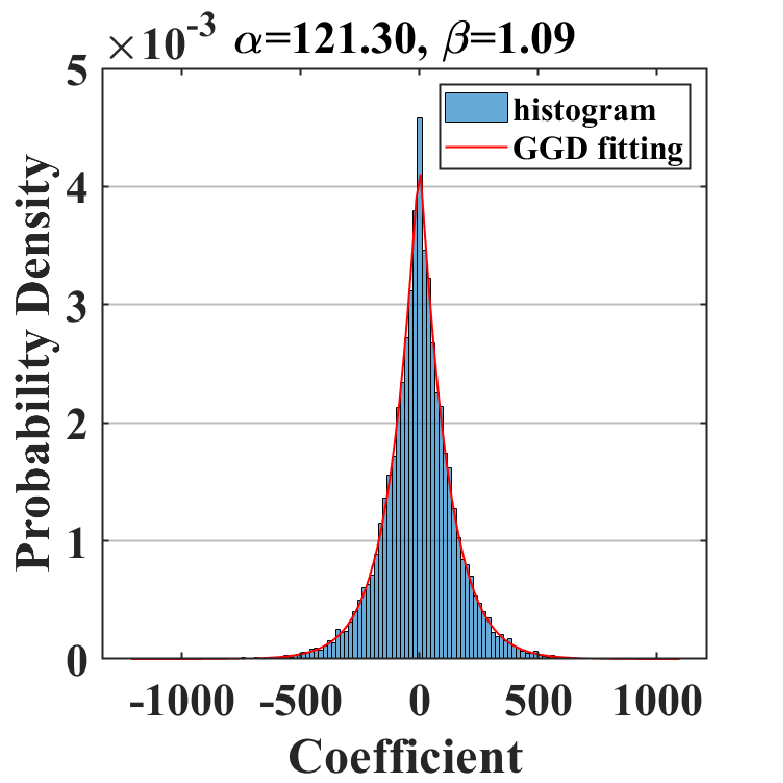}
  \subcaption{\scriptsize $\mathbf{X}^{\text{nHL}}$ (PSNR=7.60, SSIM=0.0193)}
\end{subfigure} 
\begin{subfigure}{0.24\linewidth}
\centering
  \includegraphics[width=0.45\textwidth]{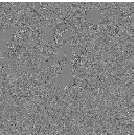}
  \hspace{0.05cm}
  \includegraphics[width=0.48\textwidth]{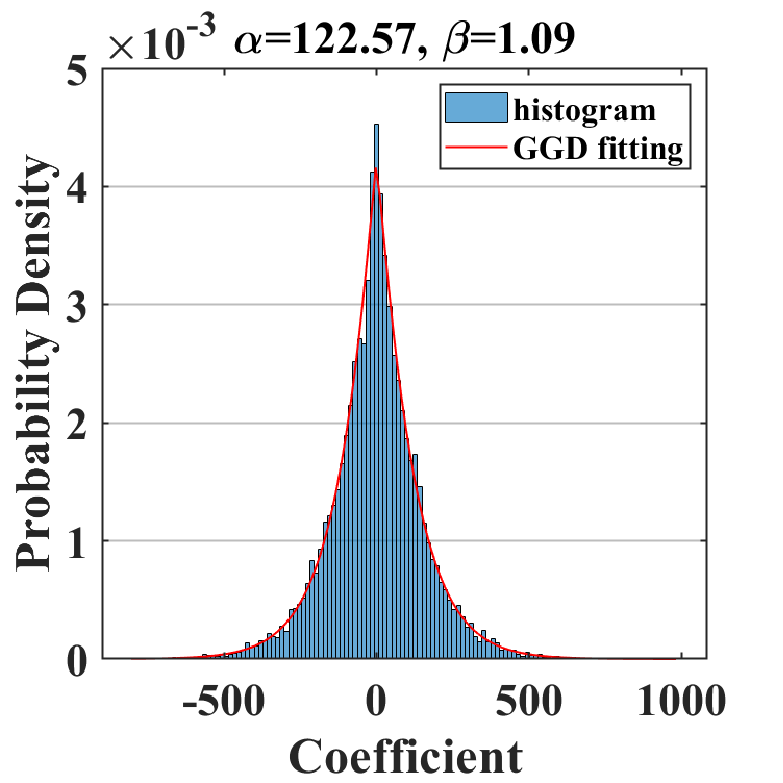}
  \subcaption{\scriptsize $\mathbf{X}^{\text{nHH}}$ (PSNR=3.40, SSIM=0.0069)}
\end{subfigure} 
\caption{Synthetic image disrupted by speckle noise ($L=1$) to study statistical property after 2D-HWT. (a) Clean SAR image and its decomposition. (b) - (g) show the speckle noise (b), noisy image (c), low-frequency sub-band $\mathbf{X}^\text{nLL}$ (d),  low-frequency sub-band $\mathbf{X}^\text{nLH}$ (e), $\mathbf{X}^\text{nHL}$ (f), $\mathbf{X}^\text{nHH}$ (g), with their histogram fitting by $\mathcal{GM}$ and $\mathcal{GGD}$ distribution, respectively. We also label the PSNR, SSIM value, and fitting parameters.}
\label{fig:wave}
\end{figure*}
\begin{theorem}
\label{the:low gamma}
Let $\mathbf{X}=\left\{x[m,n]\right\}$ be an image of size $2^J\times 2^J$, where $x[m,n]$ are independent and identically distributed random variables following the Gamma distribution $\mathcal{GM}(a, b)$. Consider the 2D multi-resolution analysis of $\mathbf{X}$ using 2D separable orthonormal Haar filters. Then the wavelet coefficients at scale $j$ exhibit the following properties:
\begin{enumerate}
    \item The low-frequency coefficients (LL sub-band) $\mathbf{A}_j(k,l)$ at scale $j$ follow a Gamma distribution: 
        \begin{equation*}
            \mathbf{A}_j(k,l) \sim \mathcal{GM}\left(2^{2j}a, 2^{-j} b\right),
        \end{equation*}
        where $1\leq k,l \leq 2^{J-j}$.
	\item The high-frequency coefficients (LH, HL, and HH sub-bands) $\mathbf{D}^i_j(k,l)$ at scale $j$ follow a symmetric distribution, where $i=\text{LH},\text{HL},\text{HH}$ and $1\leq k,l \leq 2^{J-j}$.
\end{enumerate}
\end{theorem}
\begin{proof}
For the 2D image, we use wavelet filters in (\ref{eq:wavelet}) to decompose it into low-frequency $\mathbf{A}_j$ and high-frequency coefficients $\mathbf{D}^i_j$ at the scale $j$ as follows:
\begin{equation*}
\medmuskip=0mu
\thickmuskip=0mu
\thinmuskip=0mu
\begin{aligned}
\mathbf{A}_j(k,l)&=\sum_{m=1}^{2}\sum_{n=1}^{2}\boldsymbol{\omega}^{\text{LL}}\left(m,n\right)\mathbf{A}_{j-1}\left(2k-m,2l-n\right),\\
\mathbf{D}^i_j(k,l)&=\sum_{m=1}^{2}\sum_{n=1}^{2}\boldsymbol{\omega}^{i}\left(m,n\right)\mathbf{A}_{j-1}\left(2k-m,2l-n\right).
\end{aligned}
\end{equation*}

At the first decomposition level, it is directly to calculate that:
\begin{equation*}
\mathbf{A}_1(k,l)=\frac{1}{2}\sum_{m=1}^{2}\sum_{n=1}^{2}x[2k-m,2l-n].
\end{equation*}
According to the properties of Gamma distribution, it follows that $\mathbf{A}_1(k,l) \sim \mathcal{GM}(2^2a, 2^{-1}b)$.\par
At the second decomposition level, we have:
\begin{equation*}
\mathbf{A}_2(k,l)=\frac{1}{2}\sum_{m=1}^{2}\sum_{n=1}^{2}\mathbf{A}_1\left(2k-m,2l-n\right).
\end{equation*}
By the same ways, one gets that $\mathbf{A}_2(k,l)\sim \mathcal{GM}(2^4a, 2^{-2}b)$. Therefore, at the $j$-th decomposition level, we conclude that $\mathbf{A}_j(k,l) \sim \mathcal{GM}(2^{2j}a, 2^{-j} b)$.\par
The high-frequency coefficient $\mathbf{D}^{\text{HH}}_1(k,l)$ is calculated as 
\begin{equation*}
\medmuskip=0mu
\thickmuskip=0mu
\thinmuskip=0mu
\begin{aligned}
\mathbf{D}^{\text{HH}}_j(k,l)=&\frac{1}{2}\left(\mathbf{A}_{j-1}\left(2k-1,2l-1\right)- \mathbf{A}_{j-1}\left(2k-2,2l-1\right)\right.\\
&\left.-\mathbf{A}_{j-1}\left(2k-1,2l-2\right)+\mathbf{A}_{j-1}\left(2k-2,2l-2\right)\right)
\end{aligned}
\end{equation*}
at the first decomposition level. As above-mentioned, one has
\begin{equation*}
\medmuskip=0mu
\thickmuskip=0mu
\thinmuskip=0mu
\frac{1}{2}\left(\mathbf{A}_{j-1}\left(2k-1,2l-1\right)+\mathbf{A}_{j-1}\left(2k-2,2l-2\right)\right)\sim \mathcal{GM}\left(2^{2j}a, 2^{-j} b\right)
\end{equation*} 
and 
\begin{equation*}
\medmuskip=0mu
\thickmuskip=0mu
\thinmuskip=0mu
\frac{1}{2}\left(\mathbf{A}_{j-1}\left(2k-2,2l-1\right)+\mathbf{A}_{j-1}\left(2k-1,2l-2\right)\right)\sim \mathcal{GM}\left(2^{2j}a, 2^{-j} b\right).
\end{equation*}
In this way, the distribution of $\mathbf{D}^{\text{HH}}_j(k,l)$ follows a Gamma difference distribution \cite{forresterGammaDifferenceDistribution2024}, which has the probability density function as follows:
\begin{equation*}
\medmuskip=0mu
\thickmuskip=0mu
\thinmuskip=0mu
f(z)=\left\{\begin{array}{ll}\frac{\tilde{b}^{2\tilde{a}}}{\Gamma(\tilde{a})^2}e^{\tilde{b}z}\int_{z}^{\infty}x^{\tilde{a}-1}(x-z)^{\tilde{a}-1}e^{-2\tilde{b}x}\text{d}x, \ z>0, \\ 
                                    \frac{\tilde{b}^{2\tilde{a}}}{\Gamma(\tilde{a})^2}e^{-\tilde{b}z}\int_{-z}^{\infty}x^{\tilde{a}-1}(x+z)^{\tilde{a}-1}e^{-2\tilde{b}x}\text{d}x, \ z<0\end{array}\right.
\end{equation*}
where $\tilde{a}=2^{2j}a$ and $\tilde{b}=2^{-j} b$. It is clearly that $f(z)$ is symmetric centered in zero. In addition, based on same derivation, we can conclude that all high-frequency coefficients $\mathbf{D}^{i}_j$ follow a symmetric distribution.\par
\end{proof}
A SAR image corrupted by speckle noise with $L=1$ is used as an illustrative example, as shown in figure \ref{fig:wave}. In this paper, we only decompose the SAR image at scale of $j=1$ for efficient implementation. When clean SAR images are decomposed via the 2D-HWT, they exhibit distinct characteristics across four different sub-spaces. However, when corrupted by speckle noise, most image structures become obscured, and the entire image can be closely approximated by a $\mathcal{GM}(0.87,167.89)$ distribution. Although the spatial statistical distributions vary, their visual manifestations remain difficult to interpret. By applying wavelet transform for frequency-domain analysis, we fully decouple the spatially mixed features, revealing statistical differences in noise, arising from different spatial physical characteristics, across different frequency sub-bands. As consistent with theorem \ref{the:low gamma}, the low-frequency region is well characterized by $\mathcal{GM}(2.50,115.26)$, while we use the generalized Gaussian distribution, denoted by $\mathcal{GGD}$, to model the high-frequency sub-bands. The probability density function of the $\mathcal{GGD}$ is given by:
\begin{equation*}
g(x;\alpha,\beta) = \frac{\beta}{2\alpha \Gamma(1/\beta)}e^{-(\frac{|x|}{\alpha})^\beta},
\end{equation*}
where $\alpha,\ \beta >0$ represents scale and shape parameter, respectively. As shown in figure \ref{fig:wave}, the high-frequency features follow $\mathcal{GGD}(122.89,1.10),\ \mathcal{GGD}(121.30,1.09),\ \mathcal{GGD}(122.57,1.09)$, respectively. These distributional differences allow clear separation of regions with distinct statistical properties, motivating the design of adaptive denoising networks tailored to different sub-bands.\par
Additionally, it is noteworthy that wavelet decomposition effectively separates noise from structural content, providing further advantages for structure preservation. Although the high-frequency components collectively follow the same type of distribution, their parameters differ. Moreover, high-frequency structures and the degree of noise contamination vary across directional sub-bands, as reflected in the differing PSNR and SSIM values presented in figure \ref{fig:wave}. To accommodate such statistical variations, we use same but unshared modules ensuring better adaptation for high-frequency processing.\par
\subsection{Overall network structures}
The proposed SAR-FAH model employs a four-branch architecture, as shown in figure \ref{fig:process}. It comprises three core components: a LFSP-ODE module, a  HFDE module, and a Cross-FRequency Enhancement (CFRE) module. The LFSP-ODE module is based on NODE framework, modelling low-frequency denoising and structural preservation controlled by an ODE to smooth noise and avoid artifacts. The HFDE module adopts an asymmetric U-shaped design inspired by U-Net \cite{ronnebergerUNetConvolutionalNetworks2015} with DeforConv, enabling multi-scale feature extraction and adaptive processing of irregular edges to preserve sharp structures and textures in high-frequency components. The CFRE consists of a residual block based on $3\times 3$ convolutional kernel and a $1\times 1$ CNN layer, which enhances the frequency feature restoration by interactiving processed low- and high-frequency components.\par
\begin{figure*}[htp]
	\centering
       \includegraphics[width=0.85\linewidth]{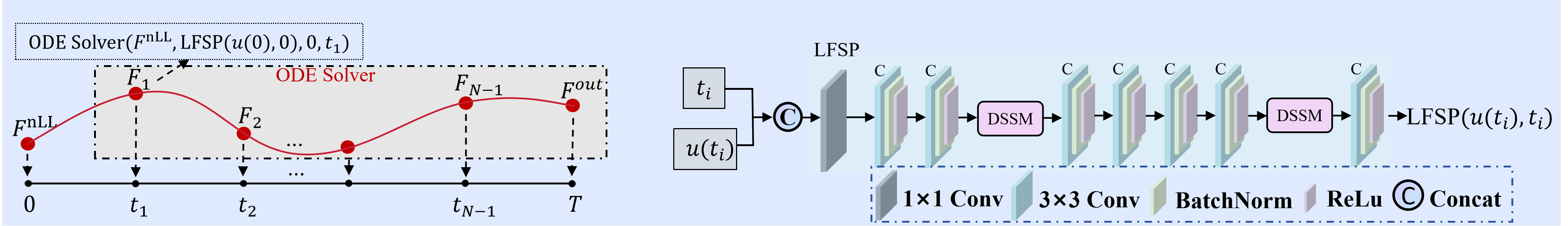}
	\caption{The structure of Low-Frequency Structural Preservation-ODE (LFSP-ODE) module. }\label{fig:LFSP}
\end{figure*}
Given a noisy SAR image $\mathbf{X}^\text{n} \in \mathbb{R}^{1 \times H \times W}$, a 2D-HWT decomposes it into one low-frequency component $\mathbf{X}^\text{nLL}$ and three high-frequency components $\left\{\mathbf{X}^\text{nLH},\ \mathbf{X}^\text{nHL},\ \mathbf{X}^\text{nHH}\right\}$ with half spatial resolution. Then a $3\times3$ convolution projects them into a higher-dimensional feature space, yielding $\mathbf{F}^\text{nLL}, \mathbf{F}^\text{nLH}, \mathbf{F}^\text{nHL}, \mathbf{F}^\text{nHH} \in \mathbb{R}^{C \times \frac{H}{2} \times \frac{W}{2}}$, with $C$ denoting the number of feature channels. The low-frequency feature $\mathbf{F}^\text{nLL}$ is processed by the LFSP-ODE module, while the high-frequency features are independently denoised and enhanced by the HFDE module, producing denoised features $\mathbf{F}^\text{LL}, \mathbf{F}^\text{LH}, \mathbf{F}^\text{HL}, \mathbf{F}^\text{HH} \in \mathbb{R}^{C \times \frac{H}{2} \times \frac{W}{2}}$. Then, these features are concatenated along the channel dimension and interacted via a CFRE module for frequency feature enhancement, resulting in a clean frequency feature. Finally, the image is reconstructed using the inverse wavelet transform.\par
Given a training set of noisy-clean image pairs $\left\{\mathbf{X}^\text{n}_i, \ \mathbf{X}_i\right\}_{i=1}^{N}$, the model is optimized using the $L_1$ loss function:
\begin{equation*}
\mathcal{L}(\Theta)
         =\frac{1}{N}\sum_{i=1}^{N}\parallel f_{\text{SAR-FAH}}(\mathbf{X}^\text{n}_i;\ \Theta)-\mathbf{X}_i \parallel _1,
\end{equation*}
where $\Theta$ denotes the learnable parameters.\par
\subsection{Low-frequency structural preservation-ODE module}
In SAR speckle reduction, it is important to note that the low-frequency part may produce artifacts due to insufficient smoothing or result in structural loss during the denoising process. In order to completely smooth out noise while preserving the complete image structure of the output image for low-frequency feature input, a high degree of similarity between the input and output is required, which can be expressed as a controllability problem of the dynamic system. And the processing at the low frequency is a sufficiently smooth problem, as there is a mapping generated by ODE that corresponds to this process \cite{eProposalMachineLearning2017,heODEInspiredNetworkDesign2019a}. In this way, we construct LFSP-ODE module, which incorporates denoising of low-frequency regions into the NODE framework \cite{chenNeuralOrdinaryDifferential2018}, allowing features to undergo continuous transformations through hidden layers under the control of an ODE. By modelling the process as a continuous dynamical system controlled by an ODE, it mitigates information loss and artifacts due to the dependence of the solution on the initial value and the regularity of the solution \cite{yanAdverSARialRobustnessDeepa}.\par 
Guided by the dynamical system perspective in equation (\ref{eq:node}), we formulate the feature denoising process for the low-frequency component as a continuous-time evolution. This process is described by the following initial value problem defined over the temporal domain $t\in (0,T]$:
\begin{equation*}
\begin{aligned}
\frac{\text{d}\mathbf{u}(t)}{\text{d}t} & =f_{\theta}^{\text{LFSP}}(\mathbf{u}(t), t), \quad t \in (0, T], \\
\mathbf{u}(0) & = \mathbf{F}^{\text{nLL}},
\end{aligned}
\end{equation*}
where $\mathbf{u}(t)$ represents the latent feature state at time $t$. To solve this problem numerically, we discretize the temporal domain and derive the update rule:
\begin{equation}
\begin{aligned}
  \mathbf{u}(t_{i+1}) &= \mathbf{u}(t_{i}) + \int_{t_{i}}^{t_{i+1}}f_{\theta}^{\text{LFSP}}(\mathbf{u}(t), t)\text{d}t,\\
  &=\text{ODESolver}(\mathbf{u}(t_{i}),\ f_{\text{LFSP}},\ t_{i},\ t_{i+1}),
 \end{aligned}
 \label{eq:node-low}
\end{equation}
with $0 = t_0 \leq t_1 \leq \cdots \leq t_{N-1} = T$ and a step size $\frac{T}{N}$. Here, $\text{ODESolver}$ is implemented using a randomized Euler method. The final state $\mathbf{u}(T)$ corresponds to the denoised low-frequency feature $\mathbf{F}^{\text{LL}}$.\par
The vector field $f_{\theta}^{\text{LFSP}}$, parameterized by the LFSP module, maps the input $\mathbf{u}(t_i)$ concatenated with $t_i$ to an output of the same dimension as $\mathbf{u}(t_i)$. The LFSP architecture comprises 7 Conv-BatchNorm(BN)-ReLU denoising blocks, augmented by two Dynamic Attentive State-Space Fusion (DASS) modules that enhance denoising through dynamic integration of global and local features using a hybrid of VMamba \cite{liuVMambaVisualState2024} and attention mechanisms  \cite{wooCBAMConvolutionalBlock2018}, which are illustrated in the next subsection.\par
The LFSP-ODE module contributes to improved structural fidelity in the denoising process. Specifically, the solution to equation (\ref{eq:node-low}) defines a continuous flow map $\Phi(\mathbf{F}^{\text{nLL}}, t)$ originating from the initial state $\mathbf{F}^{\text{nLL}}$. Owing to the properties of $f_{\theta}^{\text{LFSP}}$, this flow is unique. As a result, $\Phi$ transforms the noisy input $\mathbf{F}^{\text{nLL}}$ into the cleaned output $\mathbf{F}^{\text{LL}}$ along a deterministic trajectory, thereby preventing structural degradation. The inherent smoothness of the mapping ensures effective noise suppression in low-frequency regions, which helps avoid artifacts in the final reconstruction. In summary, by leveraging the NODE framework, the LFSP-ODE approach effectively removes noise in low-frequency domains while adaptively preserving structural details.\par
\begin{figure*}[htp]
	\centering
       \includegraphics[width=0.8\linewidth]{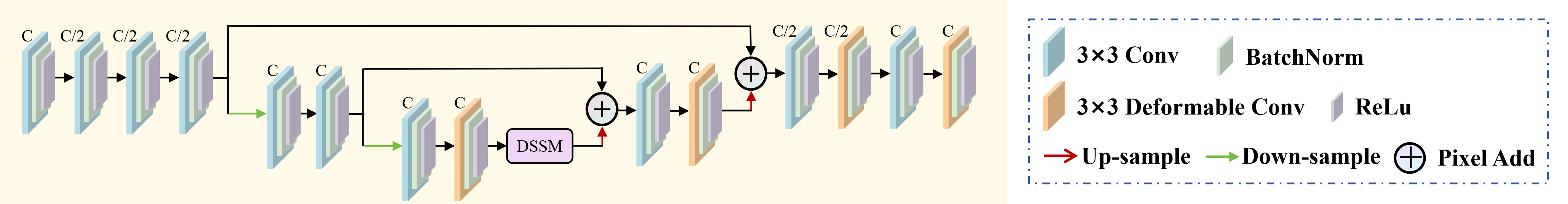}
	\caption{The structure of High-Frequency Denoising and Enhancement (HFDE) module. }\label{fig:HFDE}
\end{figure*}
\subsection{High-frequency denoising and enhancement module}
\begin{figure*}[htp]
	\centering
       \includegraphics[width=0.8\linewidth]{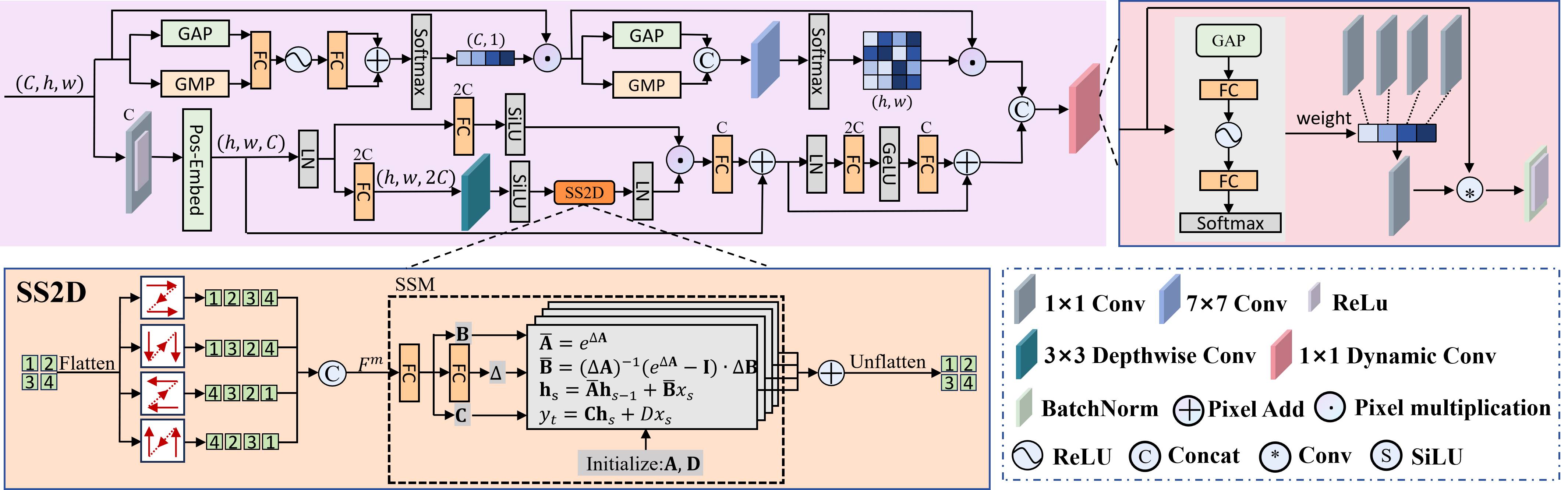}
	\caption{The structure of Dynamic Attentive State-Space fusion (DASS) module.}\label{fig:dssm}
\end{figure*}
To reduce noise in the high-frequency components $\{\mathbf{F}^\text{nLH},\mathbf{F}^\text{nHL},\mathbf{F}^\text{nHH}\}$, which contain significant sharp edges, fine textures, and strongly scattered isolated point targets, a HFDE module is designed to effectively enhance these structures. And based on difference between these components, we utilize same but unshared HFDE for separate processing. The HFDE adopts an asymmetric U-shaped architecture inspired by U-Net \cite{ronnebergerUNetConvolutionalNetworks2015}, comprising the encoder, decoder, and bottleneck parts.\par
Taking the high-frequency component $\mathbf{F}^\text{nHH}$ as an example, the process begins by feeding it into a CNN-BN-ReLU block for initial filtering and channel attention squeezing, yielding feature $\mathbf{F}^{\mathcal{E}}_0\in \mathbb{R}^{\frac{C}{2} \times \frac{H}{2} \times \frac{W}{2}}$. It is then encoded by 2  CNN-BN-ReLU-MaxPool layers with downsampling ratio of 2 as follows:
\begin{equation*}
\mathbf{F}_i^{\mathcal{E}} = \text{Encoder}(\mathbf{F}_{i-1}^{\mathcal{E}}), \quad i = 1, 2.
\end{equation*}

In the bottleneck stage, it consisted of a CNN-BN-ReLU layer and a DeforConv-BN-ReLU layer. Besides, we introduce a DASS module to extract global features and emphasize salient local regions:
\begin{equation*}
\mathbf{F}_0^{\mathcal{D}} = \text{DASS}(\mathbf{F}_2^{\mathcal{E}}).
\end{equation*}

The decoder adopts an different structures from the encoder block. Since the fixed receptive field of traditional CNNs limits their ability to capture complex geometric structures, such as multi-scale shapes and irregular boundaries, we employ deformable convolutions. These allow the network to dynamically adapt the sampling locations of the convolution kernel via learned offsets, enhancing adaptability to geometric variations. 
The DeforConv operations on a feature map $\mathbf{X}$ is defined as:
\begin{equation*}
\mathbf{Y}(k_0) = \sum_{k\in \mathcal{R}}\mathbf{W}(k_0) \cdot \mathbf{X}(k_0+k+\Delta k),
\end{equation*}
where $k_0$ is the spatial location, $\mathcal{R}$ is sampling the regular grid of convolution, $\mathbf{W}$ is kernel weight, and $\Delta k$ denotes the learned offset. The offset is learned from the preceding feature maps via additional convolutional layers. Thus, it can adaptively capture the vary of different feature maps. 
In addition, dense skip connections are incorporated to mitigate information loss. The decoding process is formulated as:
\begin{equation*}
	\mathbf{F}_{i}^{\mathcal{D}} = \text{Decoder}(\mathbf{F}_{i-1}^{\mathcal{E}}+\mathbf{F}_{i-1}^{\mathcal{D}}).
\end{equation*}
Finally, two CNN-BN-ReLU with kernel size $3\times 3$ are used to obtain the high-frequency denoising output $\mathbf{F}^\text{HH}\in \mathbb{R}^{C \times \frac{H}{2} \times \frac{W}{2}}$.\par
\subsection{Dynamic attentive state-space fusion module}
To effectively preserve structural information, it is essential to extract rich features from both low-frequency and high-frequency components of an image. To this end, we design a DASS module that adaptively extracts both local and global features for denoising. As illustrated in figure \ref{fig:dssm}, DASS adopts a dual-branch architecture that integrates a local detailed context branch and a global context branch, leveraging their complementary strengths through adaptive fusion. The detail branch incorporates a CBAM block \cite{wooCBAMConvolutionalBlock2018}, which enhances useful features and structural details through sequential channel and spatial attention. The global branch employs a Mamba-based structure to efficiently capture global contextual information with linear computational complexity.\par
Given an intermediate feature map $\mathbf{F}^{\text{m}} \in \mathbb{R}^{C \times H \times W}$, the detail branch generates a channel attention map $\mathbf{W}^{\text{cha}} \in \mathbb{R}^{C \times 1 \times 1}$ and a spatial attention map $\mathbf{W}^{\text{spa}} \in \mathbb{R}^{1 \times H \times W}$ through pooling operations to highlights informative channels and suppresses noise and emphasize spatially important regions, respectively. The output of this branch is computed as:
\begin{equation*}
\mathbf{F}^{\text{at}} = \mathbf{W}^{\text{spa}} \odot \left( \mathbf{W}^{\text{cha}} \odot \mathbf{F}^{\text{m}} \right),
\end{equation*}
where $\odot$ denotes element-wise multiplication.\par
Global feature representation is crucial for reconstructing image structures, yet CNNs are inherently limited by local receptive fields. To overcome this, we integrate a visual state space (VSS) block from VMamba \cite{liuVMambaVisualState2024} to efficiently model long-range dependencies. The process is formulated as:
\begin{equation*}
\begin{aligned}
\tilde{\mathbf{F}}^{\text{m}} &= \text{LN}(\text{Conv}_{1\times1}\left(\mathbf{F}^\text{m}) + \text{Pos-Embed}\right), \\
\tilde{\mathbf{F}}^\text{vss}_1 &= \text{SiLU}(\text{FC}(\tilde{\mathbf{F}}^{\text{m}})),\\
\tilde{\mathbf{F}}^\text{vss}_2 &= \text{LN}\left( \text{SS2D}\left( \sigma(\text{FC}(\text{DWConv}(\tilde{\mathbf{F}}^{\text{m}}))) \right) \right),\\
\tilde{\mathbf{F}}^\text{vss} &= (\tilde{\mathbf{F}}^\text{vss}_1  \odot \tilde{\mathbf{F}}^\text{vss}_1 ) + \tilde{\mathbf{F}}^{\text{m}},\\
\mathbf{F}^\text{vss}& = \tilde{\mathbf{F}}^\text{vss} + \text{FFN}(\tilde{\mathbf{F}}^\text{vss}).
\end{aligned}
\end{equation*}
where LN denotes LayerNorm, and FFN is a feed-forward network composed of two fully-connected (FC) layers and a Gaussian error linear unit. The SS2D mechanism transforms 2D features into sequences through four directional scans, captures global dependencies using SSM according to (\ref{eq:ssm}), and then fused through weighted integration to reconstruct a 2D feature map.\par
Finally, the local detail feature $\mathbf{F}^{\text{at}}$ and global feature $\mathbf{F}^{\text{vss}}$ are adaptively fused using dynamic convolution (DConv) \cite{chenDynamicConvolutionAttention2020a}. The features are first concatenated to form $\mathbf{F}^{\text{f}} \in \mathbb{R}^{2C \times H \times W}$. A soft attention vector $\boldsymbol{\alpha} \in \mathbb{R}^K$ is generated through global average pooling and fully connected layers:
\begin{equation*}
\boldsymbol{\alpha} = \text{Softmax}\left( \text{FC}\left( \text{ReLU}\left( \text{FC}\left( \text{GAP}(\mathbf{F}^{\text{f}}) \right) \right) \right) \right),
\end{equation*}
which determines the weights for $K$ convolutional kernels of size $k \times k$. The dynamic kernel and bias are computed as:
\begin{equation*}
\mathbf{W} = \sum_{i=1}^{K} \boldsymbol{\alpha}_i \tilde{\mathbf{W}}_i, \ \mathbf{b}= \sum_{i=1}^{K}\boldsymbol{\alpha}_i \tilde{\mathbf{b}}_i.
\end{equation*}
The concatenated feature is then fused as:
\begin{equation*}
\tilde{\mathbf{F}}^{\text{f}} = \text{ReLU}\left( \text{BN}\left( \mathbf{W} * \mathbf{F}^{\text{f}} + \mathbf{b} \right) \right),
\end{equation*}
where $*$ denotes the convolution operation.\par
The overall function of the DASS module can be summarized as:
\begin{equation*}
\mathbf{F}_0^{\mathcal{D}} = \text{DConv}_{1\times1} \left( [\mathbf{F}^{\text{at}} || \mathbf{F}^{\text{vss}}] \right),
\end{equation*}
where $\text{DConv}_{1\times1}(\cdot)$ denotes dynamic convolution with a $1\times1$ kernel, and $[\cdot||\cdot]$ represents concatenation along the channel dimension.\par
\section{Experiments and results}
\label{sec:experiment}
\subsection{Experimental Details} 
\subsubsection{Datasets}
To evaluate the proposed model, experiments are conducted on both synthetic and real SAR images. For synthetic evaluations, we use the UC Merced land-use (UCL) \cite{yangBagofvisualwordsSpatialExtensions2010} dataset, which consists of images from the USGS National Map Urban Area Imagery. It includes $21$ land-use categories, each containing $100$ images. From each category, $70$ images are randomly selected for training, $15$ for validation, and $1$ for testing. To further assess texture preservation performance, we also experiment on two texture datasets: the Describable Textures Dataset (DTD) \cite{cimpoiDescribingTexturesWild2014} and the Flickr Material Database (FMD) \cite{sharanAccuracySpeedMaterial2014}. The DTD contains 5,640 texture images categorized into 47 human-defined attributes. The FMD includes 1,000 images across 10 material categories. From each category in both DTD and FMD, we select 30 images for training, 10 for validation, and 1 for testing. All images are cropped to $256 \times 256$ for consistency.\par
\begin{figure}[!h]
\centering
\begin{subfigure}[b]{0.12\textwidth}
\includegraphics[width=\textwidth]{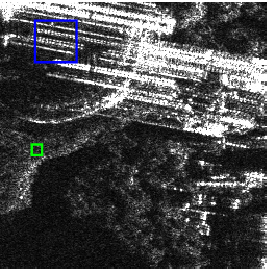}
\subcaption{GaoFen-L1}
\label{fig:SAR1}
\end{subfigure}
\begin{subfigure}[b]{0.12\textwidth}
        \includegraphics[width=\textwidth]{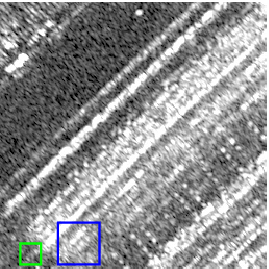}
        \subcaption{Capella-L4}
		\label{fig:SAR2}
\end{subfigure}
\begin{subfigure}[b]{0.12\textwidth}
        \includegraphics[width=\textwidth]{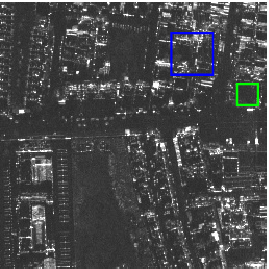}
        \subcaption{Iceye-L10}
		\label{fig:SAR3}
\end{subfigure}
\caption{Real-SAR testing images.}
\label{fig:realimg}
\end{figure}

For real SAR image evaluation, we test the proposed SAR-FAH model on three real SAR images acquired by different sensors and with varying numbers of looks, as shown in figure~\ref{fig:realimg}. Specifically, figure~\ref{fig:SAR1} shows a single-look image from the GaoFen-1 satellite (Air-SAR dataset, 3-m resolution), from which a $256 \times 256$ region is cropped. Figure~\ref{fig:SAR2} is a four-look image from the Capella satellite. Figure~\ref{fig:SAR3} is an Iceye SAR fine-mode image covering Doha International Airport, Qatar, acquired on April 29, 2024.\par
\subsubsection{Training setting} 
The images are divided into $128 \times 128$ patches and augmented for training and validation, resulting in a total of $29,400$ patches. The model is trained for 20 epochs using the Adam optimizer. The learning rate is set to $1e-3$ that would be reduced to $1e-6$ with the cosine annealing. The feature channel $C$ is set to 128. The torchdiffeq\footnote{\url{https://github.com/rtqichen/torchdiffeq}} package is employed to solve the NODE with parameters $T = 1$ and $N = 4$. All experiments are implemented using PyTorch 2.2.2 under Python 3.10, and executed on a NVIDIA GeForce GTX 3090 GPU with 24 GB of RAM.\par
\subsubsection{Compared Methods and Quality indices} 
To assess the despeckling performance, we compare the SAR-FAH model with several existing methods: POTDF \cite{xuPatchOrderingBasedSAR2015}, MuLoG \cite{deledalleMuLoGHowApply2017}, SAR-CNN \cite{chierchiaSARImageDespeckling2017}, SAR-CAM \cite{koSARImageDespeckling2022}, AGSDNet \cite{thakurAGSDNetAttentionGradientBased2022}, SAR-TRN \cite{pereraTransformerBasedSARImage2022}, and HTNet \cite{chengTwoStreamMultiplicativeHeavyTail2023}. For deep-learning methods, we train on same training dataset and use hyper-parameters as initial lectures.\par
To evalute despecking results on synthetic experiments, we employ five quality indices (QIs), including peak signal-to-noise ratio (PSNR), structural similarity (SSIM), mean average error (MAE), gradient-based structural similarity (GSSIM) \cite{chenGradientBasedStructuralSimilarity2006}, and image information correlation coefficient (IICC) \cite{wangWhyImageQuality2002}. 
Higher PSNR, SSIM, GSSIM, IICC, and lower MAE indicates better performance.\par
The quantitative evaluation results are listed in table \ref{tab:SYNTHETIC}, \ref{tab:textural}, and \ref{tab:REAL}, with the best and second-best performances highlighted in red and blue, respectively.\par
\subsection{Results on synthetic UCL dataset}
Table \ref{tab:SYNTHETIC} presents the quantitative evaluation results on the UCL dataset. The proposed SAR-FAH model achieves the highest QIs across all noise levels.\par
\begin{figure*}
\centering
\setlength{\abovecaptionskip}{0cm}
\begin{minipage}[c]{0.01\textwidth}
\centering
\rotatebox{90}{Farmland}
\end{minipage}
\begin{minipage}[c]{0.98\textwidth}
\begin{subfigure}[b]{0.09\textwidth}
\includegraphics[width=\textwidth]{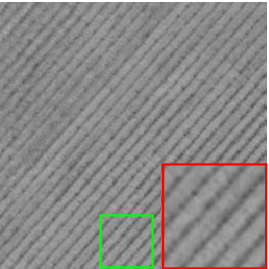}
\subcaption*{}
\end{subfigure}
\begin{subfigure}[b]{0.09\textwidth}
\includegraphics[width=\textwidth]{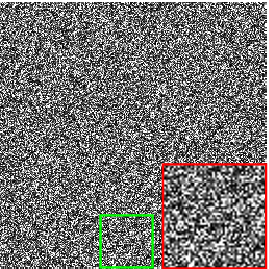}
\subcaption*{}
\end{subfigure}
\begin{subfigure}[b]{0.09\textwidth}
\includegraphics[width=\textwidth]{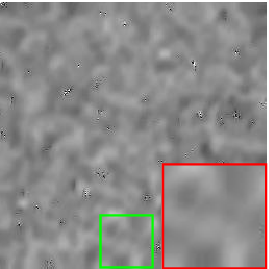}
\subcaption*{}
\end{subfigure}
\begin{subfigure}[b]{0.09\textwidth}
\includegraphics[width=\textwidth]{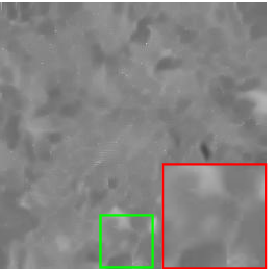}
\subcaption*{}
\end{subfigure}
\begin{subfigure}[b]{0.09\textwidth}
\includegraphics[width=\textwidth]{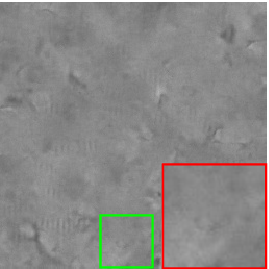}
\subcaption*{}
\end{subfigure}
\begin{subfigure}[b]{0.09\textwidth}
\includegraphics[width=\textwidth]{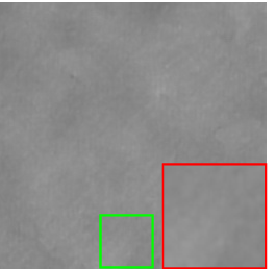}
\subcaption*{}
\end{subfigure}
\begin{subfigure}[b]{0.09\textwidth}
\includegraphics[width=\textwidth]{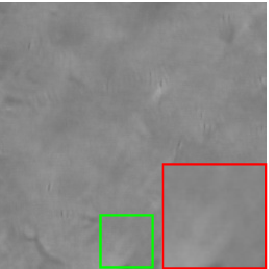}
\subcaption*{}
\end{subfigure}
\begin{subfigure}[b]{0.09\textwidth}
\includegraphics[width=\textwidth]{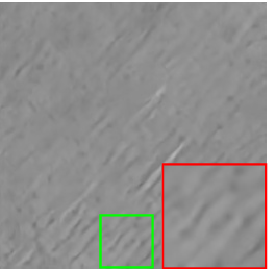}
\subcaption*{}
\end{subfigure}
\begin{subfigure}[b]{0.09\textwidth}
\includegraphics[width=\textwidth]{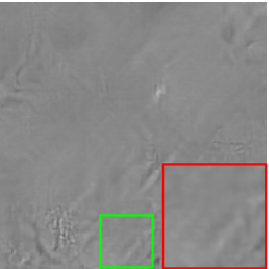}
\subcaption*{}
\end{subfigure}
\begin{subfigure}[b]{0.09\textwidth}
\includegraphics[width=\textwidth]{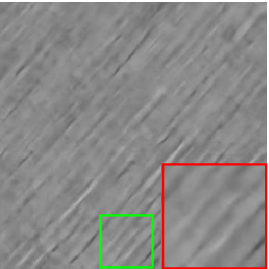}
\subcaption*{}
\end{subfigure}
\end{minipage}
\vspace{-0.4cm}
\\
\begin{minipage}[c]{0.01\textwidth}
\centering
\rotatebox{90}{Runways}
\end{minipage}
\begin{minipage}[c]{0.98\textwidth}
\begin{subfigure}[b]{0.09\textwidth}
\includegraphics[width=\textwidth]{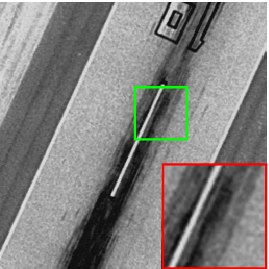}
\subcaption*{}
\end{subfigure}
\begin{subfigure}[b]{0.09\textwidth}
\includegraphics[width=\textwidth]{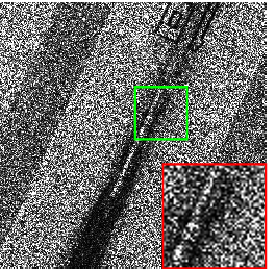}
\subcaption*{}
\end{subfigure}
\begin{subfigure}[b]{0.09\textwidth}
\includegraphics[width=\textwidth]{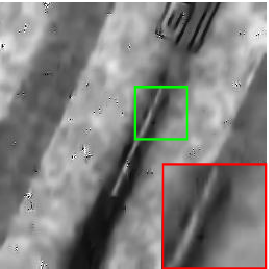}
\subcaption*{}
\end{subfigure}
\begin{subfigure}[b]{0.09\textwidth}
\includegraphics[width=\textwidth]{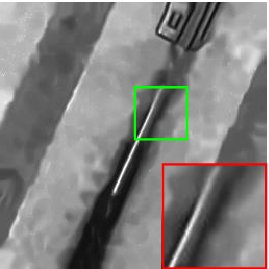}
\subcaption*{}
\end{subfigure}
\begin{subfigure}[b]{0.09\textwidth}
\includegraphics[width=\textwidth]{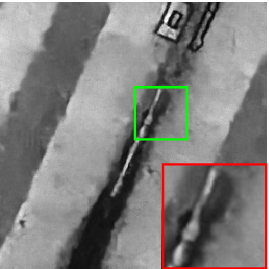}
\subcaption*{}
\end{subfigure}
\begin{subfigure}[b]{0.09\textwidth}
\includegraphics[width=\textwidth]{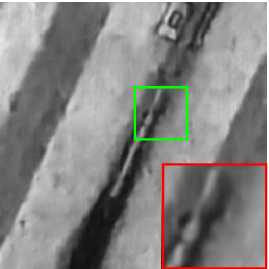}
\subcaption*{}
\end{subfigure}
\begin{subfigure}[b]{0.09\textwidth}
\includegraphics[width=\textwidth]{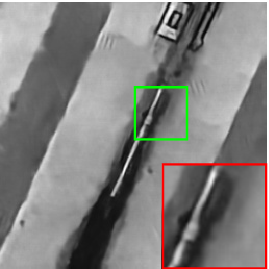}
\subcaption*{}
\end{subfigure}
\begin{subfigure}[b]{0.09\textwidth}
\includegraphics[width=\textwidth]{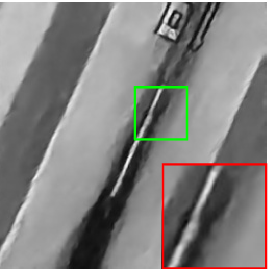}
\subcaption*{}
\end{subfigure}
\begin{subfigure}[b]{0.09\textwidth}
\includegraphics[width=\textwidth]{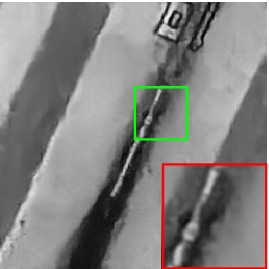}
\subcaption*{}
\end{subfigure}
\begin{subfigure}[b]{0.09\textwidth}
\includegraphics[width=\textwidth]{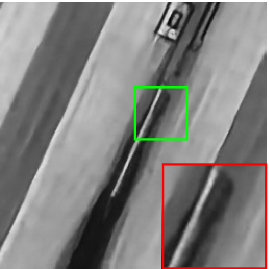}
\subcaption*{}
\end{subfigure}
\end{minipage}
\vspace{-0.4cm}
\\
\begin{minipage}[c]{0.01\textwidth}
\centering
\rotatebox{90}{Chaparral}
\end{minipage}
\begin{minipage}[c]{0.98\textwidth}
\begin{subfigure}[b]{0.09\textwidth}
\includegraphics[width=\textwidth]{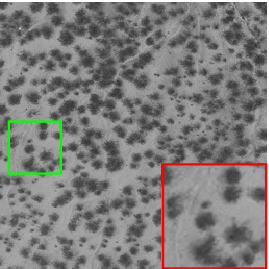}
\subcaption*{Clean}
\end{subfigure}
\begin{subfigure}[b]{0.09\textwidth}
\includegraphics[width=\textwidth]{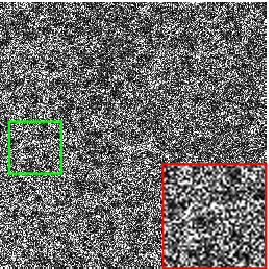}
\subcaption*{Noisy}
\end{subfigure}
\begin{subfigure}[b]{0.09\textwidth}
\includegraphics[width=\textwidth]{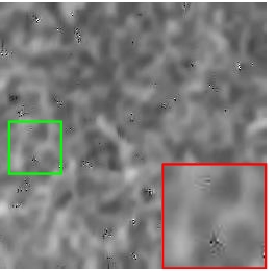}
\subcaption*{POTDF}
\end{subfigure}
\begin{subfigure}[b]{0.09\textwidth}
\includegraphics[width=\textwidth]{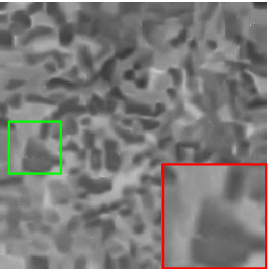}
\subcaption*{MuLoG}
\end{subfigure}
\begin{subfigure}[b]{0.09\textwidth}
\includegraphics[width=\textwidth]{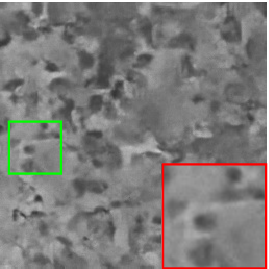}
\subcaption*{SAR-CNN}
\end{subfigure}
\begin{subfigure}[b]{0.09\textwidth}
\includegraphics[width=\textwidth]{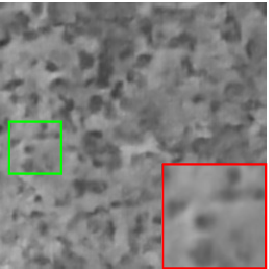}
\subcaption*{SAR-CAM}
\end{subfigure}
\begin{subfigure}[b]{0.09\textwidth}
\includegraphics[width=\textwidth]{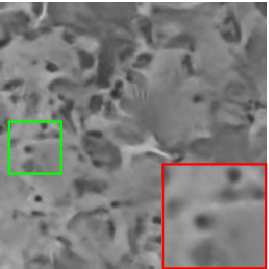}
\subcaption*{AGSDNet}
\end{subfigure}
\begin{subfigure}[b]{0.09\textwidth}
\includegraphics[width=\textwidth]{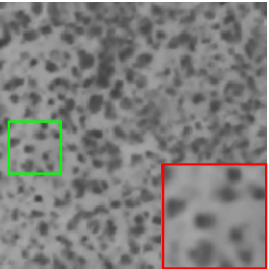}
\subcaption*{SAR-TRN}
\end{subfigure}
\begin{subfigure}[b]{0.09\textwidth}
\includegraphics[width=\textwidth]{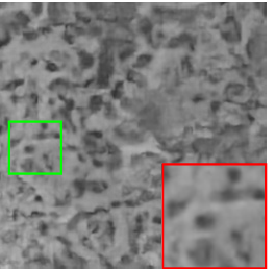}
\subcaption*{HTNet}
\end{subfigure}
\begin{subfigure}[b]{0.09\textwidth}
\includegraphics[width=\textwidth]{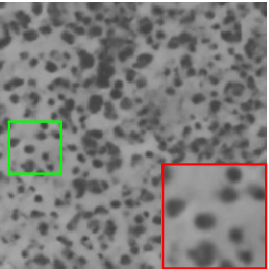}
\subcaption*{SAR-FAH}
\end{subfigure}
\end{minipage}
\label{fig:UCLL1}
\caption{Despeckling visual results on UCL dataset under noise level of $L=1$.}
\label{fig:ucl l1}
\end{figure*}

\begin{table}[hbt]
\setlength\tabcolsep{3pt}
	\centering
	\caption{The comparisons of QIs for the despeckling performance on 21 synthetic images.}\label{tab:SYNTHETIC}
	\begin{tabular}{llccccc}
		\hline
		L&Method&\multicolumn{1}{l}{PSNR}$\uparrow$ &\multicolumn{1}{l}{SSIM}$\uparrow$ &\multicolumn{1}{l}{MAE}$\downarrow$ &GSSIM$\uparrow$&IICC$\uparrow$\\
		\hline
\multirow{8}{*}{1}
&\multicolumn{1}{l}{POTDF~\cite{xuPatchOrderingBasedSAR2015}} &20.52&0.4953&17.6508 &0.4465 &0.7792 \\ 
&\multicolumn{1}{l}{MuLoG~\cite{deledalleMuLoGHowApply2017}} &22.07&0.5686&14.8156&0.4991&0.8208\\
&\multicolumn{1}{l}{SAR-CNN~\cite{chierchiaSARImageDespeckling2017}} &22.98&0.5686&13.4299&0.5367&0.8423\\
&\multicolumn{1}{l}{SAR-CAM~\cite{koSARImageDespeckling2022}} &22.68&0.5622&14.1908&0.5131&0.8356\\
&\multicolumn{1}{l}{AGSDNet~\cite{luGradDTGradientGuidedDespeckling2023}} &23.08&0.5764&13.2322&0.5276&0.8392\\
&\multicolumn{1}{l}{SAR-TRN~\cite{pereraTransformerBasedSARImage2022}}  &\color{blue}{23.37}&\color{blue}{0.6076}&\color{blue}{12.7642}&\color{blue}{0.5409}&\color{blue}{0.8620}\\
&\multicolumn{1}{l}{HTNet~\cite{chengTwoStreamMultiplicativeHeavyTail2023}} &23.29&0.5948&12.9315&0.5331&0.8505\\
&\multicolumn{1}{l}{SAR-FAH}&\color{red}{23.61}&\color{red}{0.6260}&\color{red}{12.2443}&\color{red}{0.5540}&\color{red}{0.8721} \\
\hline
\multirow{8}{*}{4}
&\multicolumn{1}{l}{POTDF~\cite{xuPatchOrderingBasedSAR2015}} &24.25 &0.6567&11.4681&0.5766&0.8862\\
&\multicolumn{1}{l}{MuLoG~\cite{deledalleMuLoGHowApply2017}}&25.17&0.6984&10.2977&0.6091&0.9101\\
&\multicolumn{1}{l}{SAR-CNN~\cite{chierchiaSARImageDespeckling2017}}&25.88&0.7060&9.4799&0.6273&0.9148\\
&\multicolumn{1}{l}{SAR-CAM~\cite{koSARImageDespeckling2022}}&25.89 &\color{blue}{0.7295}&\color{blue}{9.1478}&\color{blue}{0.6372}&0.9254\\
&\multicolumn{1}{l}{AGSDNet~\cite{luGradDTGradientGuidedDespeckling2023}}&25.86&0.7024 &9.4675 &0.6194 &0.9117 \\
&\multicolumn{1}{l}{SAR-TRN~\cite{pereraTransformerBasedSARImage2022}}&25.94&0.7238&9.4592&0.6306&\color{blue}{0.9282}\\
&\multicolumn{1}{l}{HTNet~\cite{chengTwoStreamMultiplicativeHeavyTail2023}} &\color{blue}{26.08}&0.7234&9.2817&0.6361&0.9259\\
&\multicolumn{1}{l}{SAR-FAH}&\color{red}{26.48}&\color{red}{0.7446}&\color{red}{8.7610}&\color{red}{0.6560}&\color{red}{0.9351} \\ 
\hline
\multirow{8}{*}{10}
&\multicolumn{1}{l}{POTDF~\cite{xuPatchOrderingBasedSAR2015}}&26.40& 0.7414&9.0798&0.6571& 0.9307\\
&\multicolumn{1}{l}{MuLoG~\cite{deledalleMuLoGHowApply2017}} &27.33&0.7758&8.0149&0.6867&0.9484\\
&\multicolumn{1}{l}{SAR-CNN~\cite{chierchiaSARImageDespeckling2017}} &28.00&0.7851&\color{blue}{7.4381}&0.7002&0.9497\\
&\multicolumn{1}{l}{SAR-CAM~\cite{koSARImageDespeckling2022}} &\color{blue}{28.10}&\color{blue}{0.7949}&7.4348&\color{blue}{0.7095}&\color{blue}{0.9555}\\
&\multicolumn{1}{l}{AGSDNet~\cite{luGradDTGradientGuidedDespeckling2023}} &27.78&0.7764&7.6112&0.6850&0.9467\\
&\multicolumn{1}{l}{SAR-TRN~\cite{pereraTransformerBasedSARImage2022}} &27.66&0.7830&7.8026&0.6857&0.9512\\
&\multicolumn{1}{l}{HTNet~\cite{chengTwoStreamMultiplicativeHeavyTail2023}} &28.00&0.7917&7.4624&0.7025&0.9526\\
&\multicolumn{1}{l}{SAR-FAH}&\color{red}{28.41}&\color{red}{0.8081}& \color{red}{7.0470}&\color{red}{0.7173} &\color{red}{0.9578} \\
        \hline
	\end{tabular}
\end{table}
\begin{table}[hbt]
\setlength\tabcolsep{3pt}
	\centering
	\caption{The comparisons of QIs for the despeckling performance on 57 synthetic textural images.}\label{tab:textural}
	\begin{tabular}{llccccc}
		\hline
		L&Method&\multicolumn{1}{l}{PSNR}$\uparrow$ &\multicolumn{1}{l}{SSIM}$\uparrow$ &\multicolumn{1}{l}{MAE}$\downarrow$ &GSSIM$\uparrow$&IICC$\uparrow$\\
		\hline
\multirow{8}{*}{1}
&\multicolumn{1}{l}{POTDF~\cite{xuPatchOrderingBasedSAR2015}} &20.49&0.4782&20.4881&0.5180&0.7414 \\ 
&\multicolumn{1}{l}{MuLoG~\cite{deledalleMuLoGHowApply2017}} &22.20&0.5582&15.3374&0.5906&0.7944 \\
&\multicolumn{1}{l}{SAR-CNN~\cite{chierchiaSARImageDespeckling2017}} &23.20&0.5579&13.8903&0.6027&0.8115\\
&\multicolumn{1}{l}{SAR-CAM~\cite{koSARImageDespeckling2022}} &23.04&0.5636&14.4586&0.5902&0.8113\\
&\multicolumn{1}{l}{AGSDNet~\cite{luGradDTGradientGuidedDespeckling2023}} &23.31&0.5651&13.8043&0.6025&0.8085\\
&\multicolumn{1}{l}{SAR-TRN~\cite{pereraTransformerBasedSARImage2022}}&\color{blue}{23.87}&\color{blue}{0.6071}&13.1174&\color{blue}{0.6154}&\color{blue}{0.8353}\\
&\multicolumn{1}{l}{HTNet~\cite{chengTwoStreamMultiplicativeHeavyTail2023}} &\color{blue}{23.87}&0.6038&\color{blue}{13.0610}&\color{blue}{0.6154}&0.8295\\
&\multicolumn{1}{l}{SAR-FAH}&\color{red}{23.98}&\color{red}{0.6146}&\color{red}{12.8600}&\color{red}{0.6191}&\color{red}{0.8359}\\
\hline
\multirow{8}{*}{4}
&\multicolumn{1}{l}{POTDF~\cite{xuPatchOrderingBasedSAR2015}} & 24.59&0.6458& 11.7017&0.6376 &0.8624\\
&\multicolumn{1}{l}{MuLoG~\cite{deledalleMuLoGHowApply2017}} &25.27&0.6859&10.8203&0.6703&0.8797\\
&\multicolumn{1}{l}{SAR-CNN~\cite{chierchiaSARImageDespeckling2017}} &25.99&0.6953&10.1294&0.6818&0.8872\\
&\multicolumn{1}{l}{SAR-CAM~\cite{koSARImageDespeckling2022}} &25.84&0.6855&10.5167&0.6693 &0.8865\\
&\multicolumn{1}{l}{AGSDNet~\cite{luGradDTGradientGuidedDespeckling2023}} &25.84&0.6833&10.3166&0.6720&0.8817\\
&\multicolumn{1}{l}{SAR-TRN~\cite{pereraTransformerBasedSARImage2022}}&25.85&0.7081&10.3923&0.6809&0.8930\\
&\multicolumn{1}{l}{HTNet~\cite{chengTwoStreamMultiplicativeHeavyTail2023}} &\color{blue}{26.34}&\color{blue}{0.7149}&\color{blue}{9.8157}&\color{blue}{0.6872}&\color{blue}{0.8940}\\
&\multicolumn{1}{l}{SAR-FAH}&\color{red}{26.59}&\color{red}{0.7209}&\color{red}{9.5759}& \color{red}{0.6912}&\color{red}{0.8987} \\
\hline
\multirow{8}{*}{10}
&\multicolumn{1}{l}{POTDF~\cite{xuPatchOrderingBasedSAR2015}} &26.78 &0.7302&9.2264&0.7008&0.9075 \\ 
&\multicolumn{1}{l}{MuLoG~\cite{deledalleMuLoGHowApply2017}} &27.39&0.7585&8.5393&0.7229&0.9162 \\
&\multicolumn{1}{l}{SAR-CNN~\cite{chierchiaSARImageDespeckling2017}} &27.74&0.7640&8.2928&0.7319&0.9195 \\
&\multicolumn{1}{l}{SAR-CAM~\cite{koSARImageDespeckling2022}} &28.01&0.7707&8.2424 &0.7262&0.9222 \\
&\multicolumn{1}{l}{AGSDNet~\cite{luGradDTGradientGuidedDespeckling2023}} &27.57&0.7509&8.5154&0.7187&0.9145 \\
&\multicolumn{1}{l}{SAR-TRN~\cite{pereraTransformerBasedSARImage2022}}&26.93&0.7629&9.2541&0.7161&0.9191 \\
&\multicolumn{1}{l}{HTNet~\cite{chengTwoStreamMultiplicativeHeavyTail2023}} &\color{blue}{28.09}&\color{blue}{0.7817}&\color{blue}{8.0362}&\color{blue}{0.7357}&\color{blue}{0.9248}\\
&\multicolumn{1}{l}{SAR-FAH}&\color{red}{28.38}&\color{red}{0.7891}&\color{red}{7.8152}&\color{red}{0.7402}&\color{red}{0.9282} \\
        \hline
	\end{tabular}
\end{table}
Visual results on UCL dataset, including farmland, runways, and chaparral, are shown in figure \ref{fig:ucl l1}. At the first row, for the farmland image with rich textures, it can be observed that SAR-TRN causes texture distortion, while other models completely blur the farmland texture. In contrast, the SAR-FAH model preserves more farmland textures without distortion. At the second row, an runways image is illustrated the edges preservations. It can be seen that the compared models not only blur the edges of the road, but also produce artifacts in homogeneous areas. However, the proposed model achieves clear edges without artifacts. In addition, for some fine textures and small isolated targets, they cannot be well restored. As shown in the third row, some seaweeds is blurred or with blurred structures, while the SAR-FAH preserves these fined objects. These better performance are attributed to the frequency-adaptive heterogeneous framework for despeckling enabling better results on both homogeneous and heterogeneous regions. Especially, the utilization of NODE framework for low-frequency processing provides better protection for image structures.\par
\subsection{Results on synthetic textural dataset}
\begin{figure*}
\setlength{\abovecaptionskip}{0cm}
\centering
\begin{subfigure}[b]{0.09\textwidth}
\includegraphics[width=\textwidth]{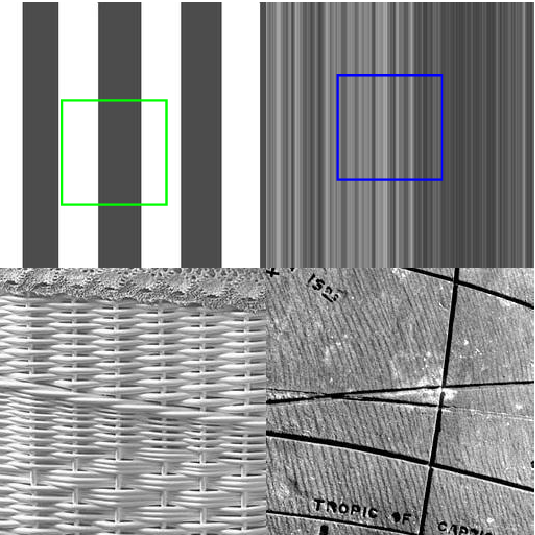}
\subcaption*{}
\end{subfigure}
\begin{subfigure}[b]{0.09\textwidth}
\includegraphics[width=\textwidth]{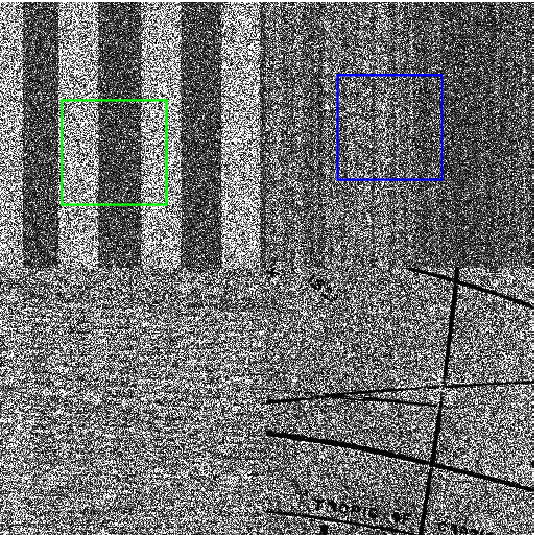}
\subcaption*{}
\end{subfigure}
\begin{subfigure}[b]{0.09\textwidth}
\includegraphics[width=\textwidth]{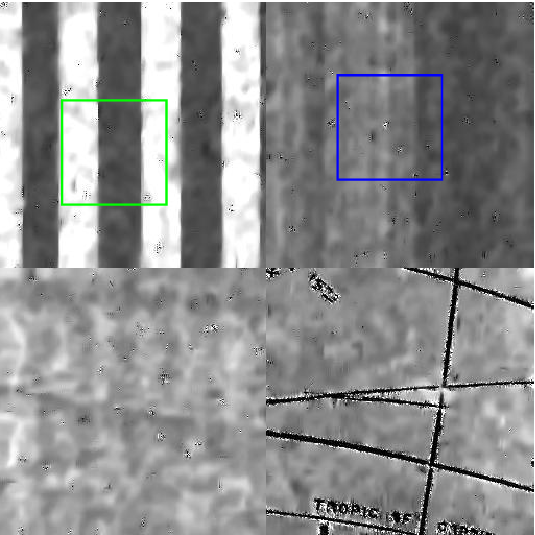}
\subcaption*{}
\end{subfigure}
\begin{subfigure}[b]{0.09\textwidth}
\includegraphics[width=\textwidth]{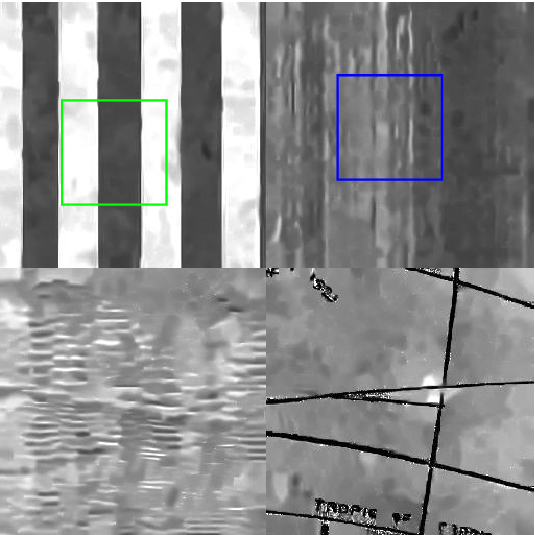}
\subcaption*{}
\end{subfigure}
\begin{subfigure}[b]{0.09\textwidth}
\includegraphics[width=\textwidth]{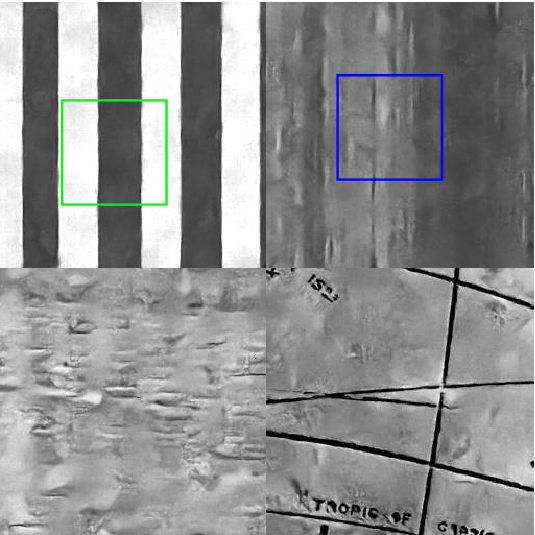}
\subcaption*{}
\end{subfigure}
\begin{subfigure}[b]{0.09\textwidth}
\includegraphics[width=\textwidth]{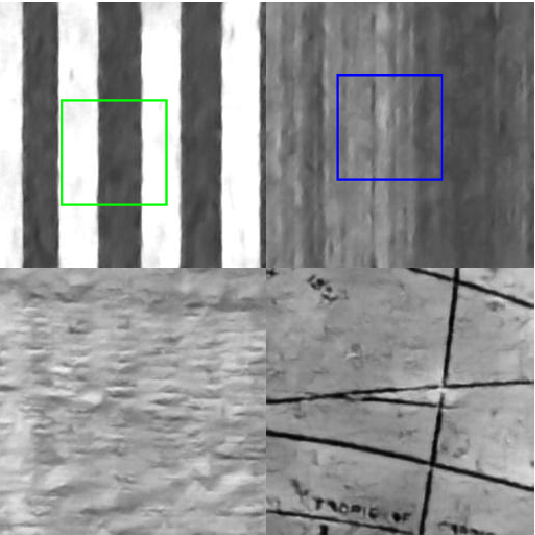}
\subcaption*{}
\end{subfigure}
\begin{subfigure}[b]{0.09\textwidth}
\includegraphics[width=\textwidth]{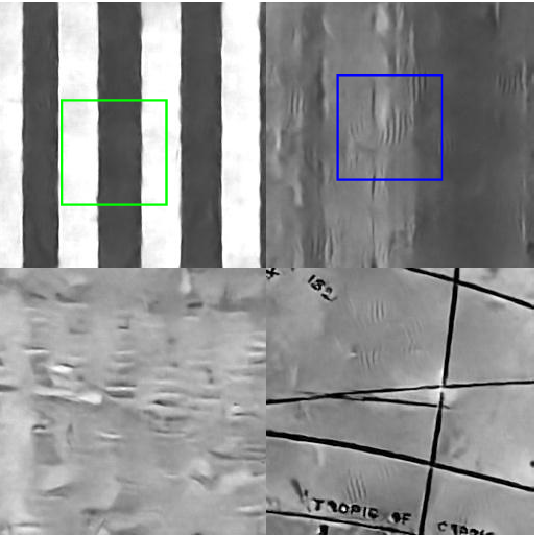}
\subcaption*{}
\end{subfigure}
\begin{subfigure}[b]{0.09\textwidth}
\includegraphics[width=\textwidth]{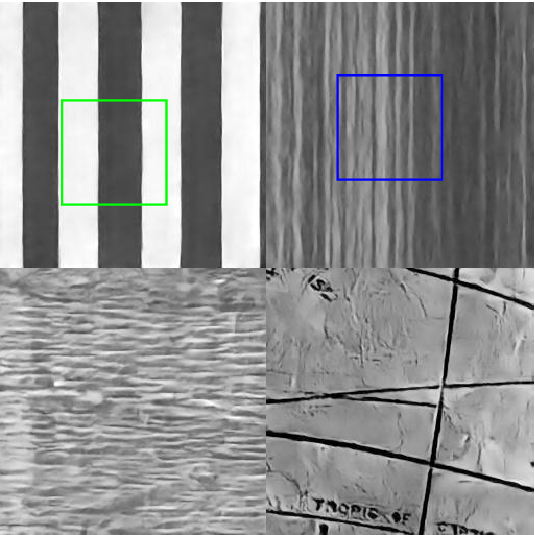}
\subcaption*{}
\end{subfigure}
\begin{subfigure}[b]{0.09\textwidth}
\includegraphics[width=\textwidth]{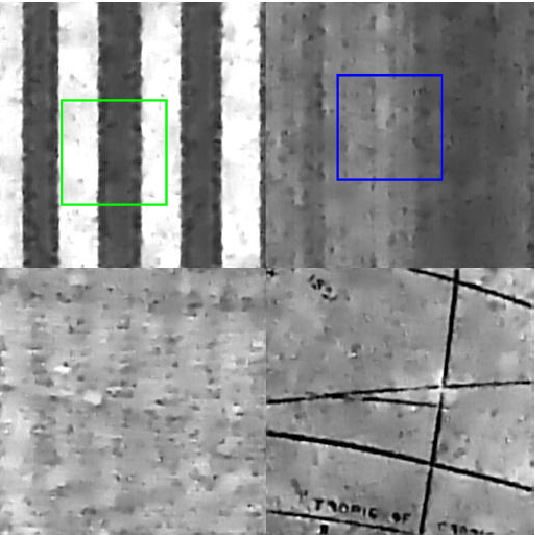}
\subcaption*{}
\end{subfigure}
\begin{subfigure}[b]{0.09\textwidth}
\includegraphics[width=\textwidth]{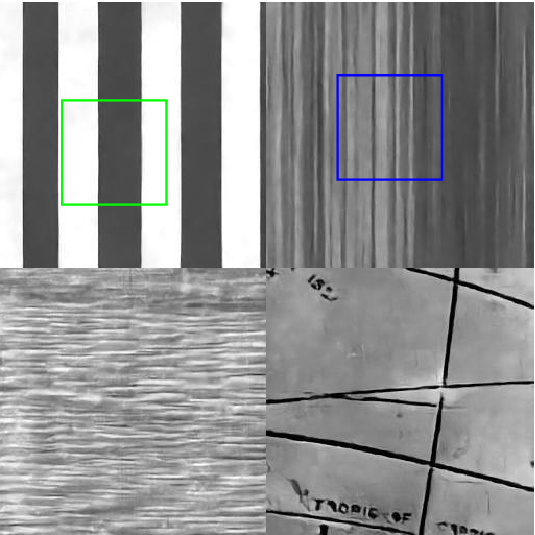}
\subcaption*{}
\end{subfigure}
\vspace{-0.4cm}
\\
\begin{subfigure}[b]{0.09\textwidth}
\includegraphics[width=\textwidth]{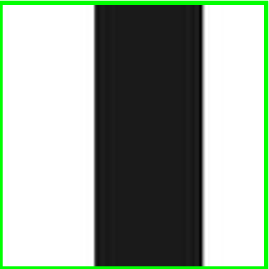}
\subcaption*{}
\end{subfigure}
\begin{subfigure}[b]{0.09\textwidth}
\includegraphics[width=\textwidth]{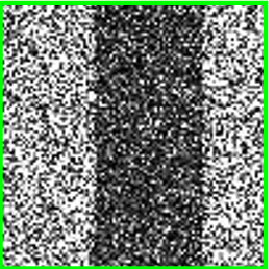}
\subcaption*{}
\end{subfigure}
\begin{subfigure}[b]{0.09\textwidth}
\includegraphics[width=\textwidth]{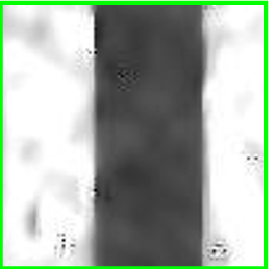}
\subcaption*{}
\end{subfigure}
\begin{subfigure}[b]{0.09\textwidth}
\includegraphics[width=\textwidth]{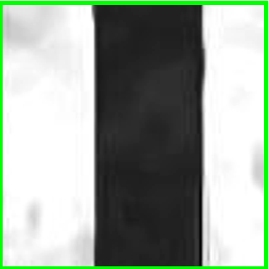}
\subcaption*{}
\end{subfigure}
\begin{subfigure}[b]{0.09\textwidth}
\includegraphics[width=\textwidth]{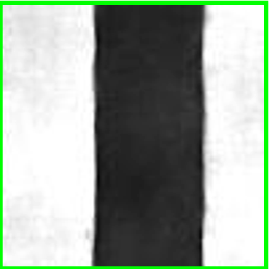}
\subcaption*{}
\end{subfigure}
\begin{subfigure}[b]{0.09\textwidth}
\includegraphics[width=\textwidth]{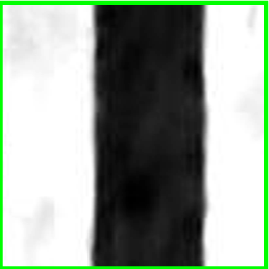}
\subcaption*{}
\end{subfigure}
\begin{subfigure}[b]{0.09\textwidth}
\includegraphics[width=\textwidth]{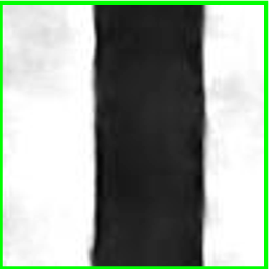}
\subcaption*{}
\end{subfigure}
\begin{subfigure}[b]{0.09\textwidth}
\includegraphics[width=\textwidth]{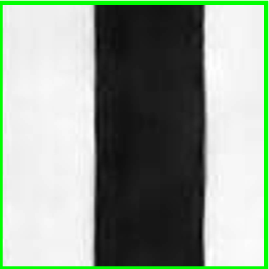}
\subcaption*{}
\end{subfigure}
\begin{subfigure}[b]{0.09\textwidth}
\includegraphics[width=\textwidth]{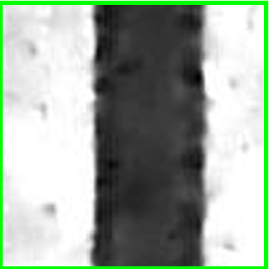}
\subcaption*{}
\end{subfigure}
\begin{subfigure}[b]{0.09\textwidth}
\includegraphics[width=\textwidth]{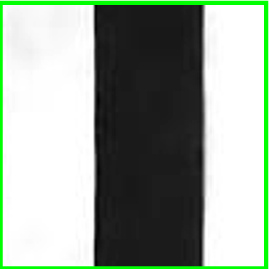}
\subcaption*{}
\end{subfigure}
\vspace{-0.4cm}
\\
\hspace{-0.23cm}
\begin{subfigure}[b]{0.09\textwidth}
\includegraphics[width=\textwidth]{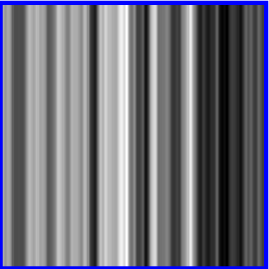}
\subcaption*{Clean}
\end{subfigure}
\begin{subfigure}[b]{0.09\textwidth}
\includegraphics[width=\textwidth]{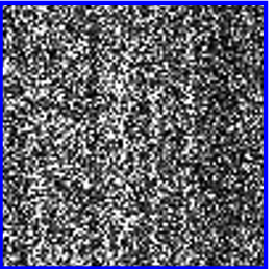}
\subcaption*{Noisy}
\end{subfigure}
\begin{subfigure}[b]{0.09\textwidth}
\includegraphics[width=\textwidth]{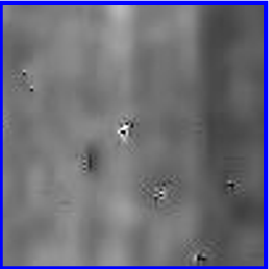}
\subcaption*{POTDF}
\end{subfigure}
\begin{subfigure}[b]{0.09\textwidth}
\includegraphics[width=\textwidth]{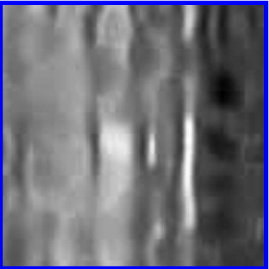}
\subcaption*{MuLoG}
\end{subfigure}
\begin{subfigure}[b]{0.09\textwidth}
\includegraphics[width=\textwidth]{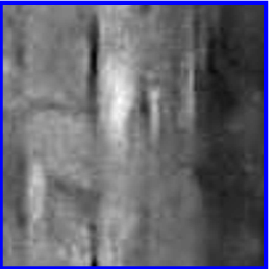}
\subcaption*{SAR-CNN}
\end{subfigure}
\begin{subfigure}[b]{0.09\textwidth}
\includegraphics[width=\textwidth]{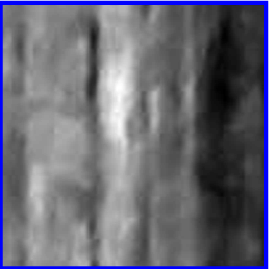}
\subcaption*{SAR-CAM}
\end{subfigure}
\begin{subfigure}[b]{0.09\textwidth}
\includegraphics[width=\textwidth]{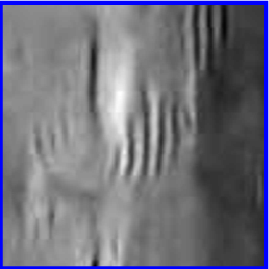}
\subcaption*{AGSDNet}
\end{subfigure}
\begin{subfigure}[b]{0.09\textwidth}
\includegraphics[width=\textwidth]{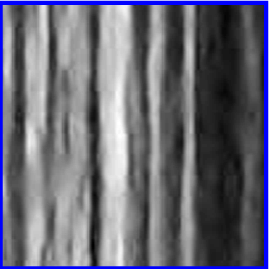}
\subcaption*{SAR-TRN}
\end{subfigure}
\begin{subfigure}[b]{0.09\textwidth}
\includegraphics[width=\textwidth]{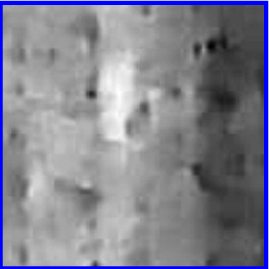}
\subcaption*{HTNet}
\end{subfigure}
\begin{subfigure}[b]{0.09\textwidth}
\includegraphics[width=\textwidth]{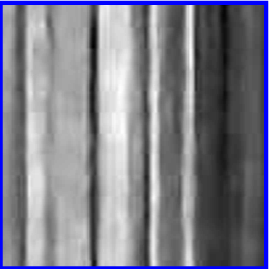}
\subcaption*{SAR-FAH}
\end{subfigure}
\caption{Despeckling visual results on textural dataset under noise level of $L=1$.}
\label{fig:tex l1}
\end{figure*} 
The proposed SAR-FAH model decomposes the image into low-frequency and high-frequency components and processes each with specialized denoising strategies to better preserve structural and textural details. To further evaluate its effectiveness in maintaining textures, a dedicated texture dataset comprising 57 categories was constructed, as summarized in table \ref{tab:textural}. Notably, SAR-FAH consistently achieves the best performance across all QIs.\par
Visual results on composite texture images, comprising banded, lined, braided, and moderate stone patterns, are presented in figure \ref{fig:tex l1}. Severe texture degradation is evident in the noisy image, especially in fine structures such as lined and braided regions, which are critical in practical applications like target boundary. These models tend to over-smooth resulting in blurred textural details. As can be seen, due to the lack of separate processing between homogeneous and heterogeneous regions, in these comparative models, both homogeneous regions and heterogeneous regions rich in texture structures have not achieved good recovery. The artifacts, texture blurring, and texture distortion phenomena are evident in the results of other models. Clearly, SAR-FAH exhibits superior texture preservation.\par
\subsection{Result on real-SAR images} 
\begin{figure*}
\setlength{\abovecaptionskip}{0cm}
\centering
\begin{subfigure}[b]{0.10\textwidth}
\includegraphics[width=\textwidth]{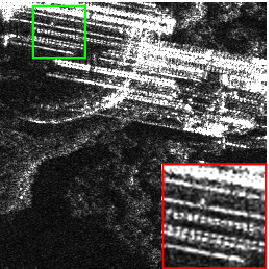}
\subcaption*{Noise}
\end{subfigure}
\begin{subfigure}[b]{0.10\textwidth}
\includegraphics[width=\textwidth]{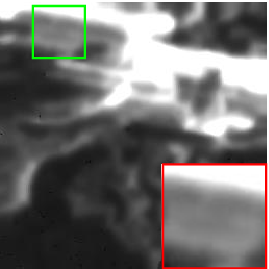}
\subcaption*{POTDF}
\end{subfigure}
\begin{subfigure}[b]{0.10\textwidth}
\includegraphics[width=\textwidth]{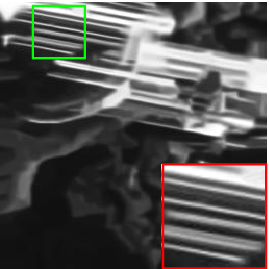}
\subcaption*{MuLoG}
\end{subfigure}
\begin{subfigure}[b]{0.10\textwidth}
\includegraphics[width=\textwidth]{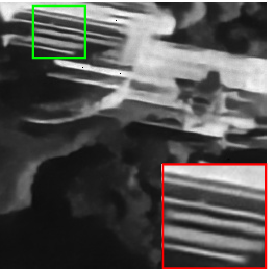}
\subcaption*{SAR-CNN}
\end{subfigure}
\begin{subfigure}[b]{0.10\textwidth}
\includegraphics[width=\textwidth]{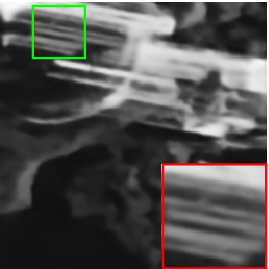}
\subcaption*{SAR-CAM}
\end{subfigure}
\begin{subfigure}[b]{0.10\textwidth}
\includegraphics[width=\textwidth]{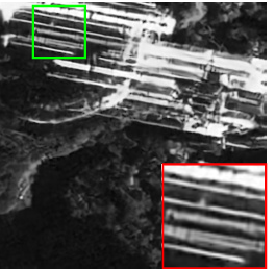}
\subcaption*{AGSDNet}
\end{subfigure}
\begin{subfigure}[b]{0.10\textwidth}
\includegraphics[width=\textwidth]{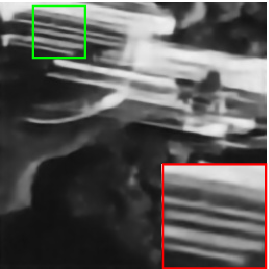}
\subcaption*{SAR-TRN}
\end{subfigure}
\begin{subfigure}[b]{0.10\textwidth}
\includegraphics[width=\textwidth]{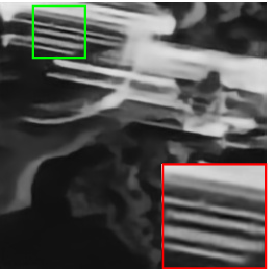}
\subcaption*{HTNet}
\end{subfigure}
\begin{subfigure}[b]{0.10\textwidth}
\includegraphics[width=\textwidth]{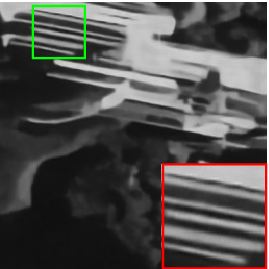}
\subcaption*{SAR-FAH}
\end{subfigure}
\caption{Despeckling visual results on GaoFen-L1 with a demarcated area zoomed in 2 times for easy comparison.}
\label{fig:real L1}
\end{figure*}

\begin{figure*}
\setlength{\abovecaptionskip}{0cm}
\centering
\begin{subfigure}[b]{0.10\textwidth}
\includegraphics[width=\textwidth]{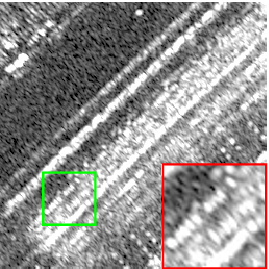}
\subcaption*{Noise}
\end{subfigure}
\begin{subfigure}[b]{0.10\textwidth}
\includegraphics[width=\textwidth]{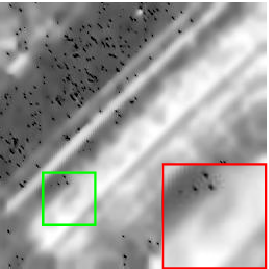}
\subcaption*{POTDF}
\end{subfigure}
\begin{subfigure}[b]{0.10\textwidth}
\includegraphics[width=\textwidth]{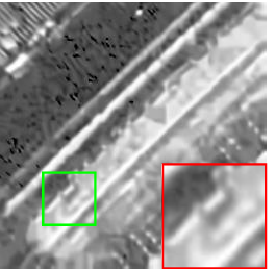}
\subcaption*{MuLoG}
\end{subfigure}
\begin{subfigure}[b]{0.10\textwidth}
\includegraphics[width=\textwidth]{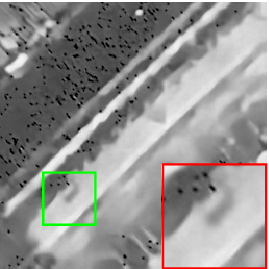}
\subcaption*{SAR-CNN}
\end{subfigure}
\begin{subfigure}[b]{0.10\textwidth}
\includegraphics[width=\textwidth]{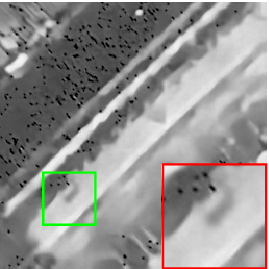}
\subcaption*{SAR-CAM}
\end{subfigure}
\begin{subfigure}[b]{0.10\textwidth}
\includegraphics[width=\textwidth]{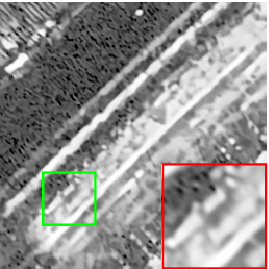}
\subcaption*{AGSDNet}
\end{subfigure}
\begin{subfigure}[b]{0.10\textwidth}
\includegraphics[width=\textwidth]{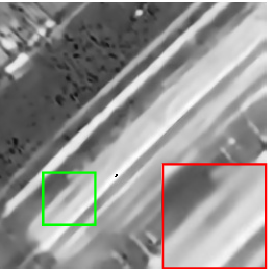}
\subcaption*{SAR-TRN}
\end{subfigure}
\begin{subfigure}[b]{0.10\textwidth}
\includegraphics[width=\textwidth]{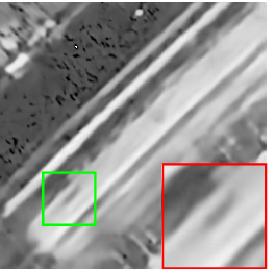}
\subcaption*{HTNet}
\end{subfigure}
\begin{subfigure}[b]{0.10\textwidth}
\includegraphics[width=\textwidth]{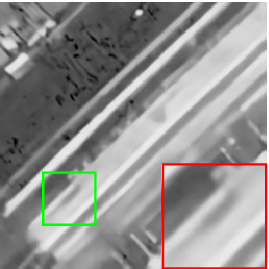}
\subcaption*{SAR-FAH}
\end{subfigure}
\caption{Despeckling visual results on Capella-L4 with a demarcated area zoomed in 2 times for easy comparison.}
\label{fig:real L4}
\end{figure*}

\begin{figure*}
\setlength{\abovecaptionskip}{0cm}
\centering
\begin{subfigure}[b]{0.10\textwidth}
\includegraphics[width=\textwidth]{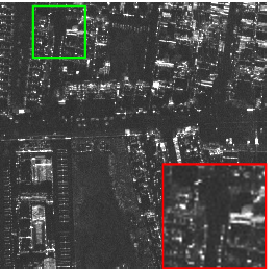}
\subcaption*{Noise}
\end{subfigure}
\begin{subfigure}[b]{0.10\textwidth}
\includegraphics[width=\textwidth]{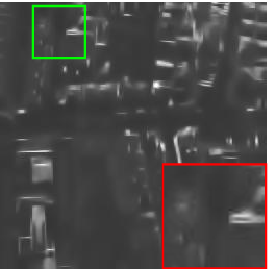}
\subcaption*{POTDF}
\end{subfigure}
\begin{subfigure}[b]{0.10\textwidth}
\includegraphics[width=\textwidth]{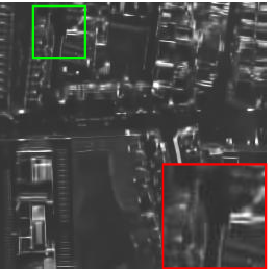}
\subcaption*{MuLoG}
\end{subfigure}
\begin{subfigure}[b]{0.10\textwidth}
\includegraphics[width=\textwidth]{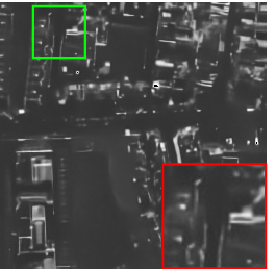}
\subcaption*{SAR-CNN}
\end{subfigure}
\begin{subfigure}[b]{0.10\textwidth}
\includegraphics[width=\textwidth]{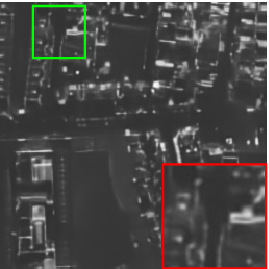}
\subcaption*{SAR-CAM}
\end{subfigure}
\begin{subfigure}[b]{0.10\textwidth}
\includegraphics[width=\textwidth]{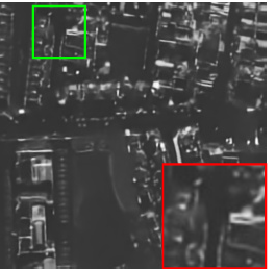}
\subcaption*{AGSDNet}
\end{subfigure}
\begin{subfigure}[b]{0.10\textwidth}
\includegraphics[width=\textwidth]{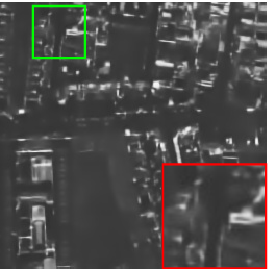}
\subcaption*{SAR-TRN}
\end{subfigure}
\begin{subfigure}[b]{0.10\textwidth}
\includegraphics[width=\textwidth]{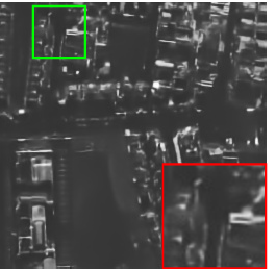}
\subcaption*{HTNet}
\end{subfigure}
\begin{subfigure}[b]{0.10\textwidth}
\includegraphics[width=\textwidth]{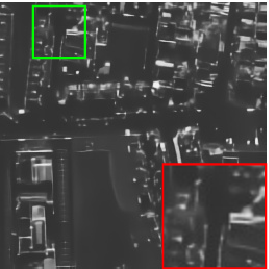}
\subcaption*{SAR-FAH}
\end{subfigure}
\caption{Despeckling visual results on Iceye-L10 with a demarcated area zoomed in 2 times for easy comparison.}
\label{fig:real L10}
\end{figure*}

\begin{table}[hbt]
	\setlength\tabcolsep{3pt}
\centering
\caption{The comparisons of no-reference QIs of the despeckling performance on 3 real-SAR images.}\label{tab:REAL}
\begin{tabular}{rrccccc}
\hline
Image&Method &ENL$\uparrow$ &MoI$\uparrow$ &
\multicolumn{2}{c}{EPD-ROA} &MoR$\uparrow$ \\ 
\cline{5-6}
&&&&HD$\uparrow$&VD$\uparrow$&\\
\hline
Best value&&$+\infty$ &1&1&1&1 \\ 
\hline
\multirow{8}{*}{GaoFen-L1}
&AGSDNet           &24.52&0.9835&0.8721&0.8454&2.2219\\ 
\cline{2-7}
&POTDF            &82.58 & 0.6242 & 0.8552 & 0.7465 & 0.6586\\
&MuLoG                 &50.03 & 0.9734 & 0.8575 & 0.7871 & 0.9593\\
&SAR-CNN                &250.62&0.9685&0.8596&\color{blue}{0.7935}&\color{blue}{0.9747}\\
&SAR-CAM             &155.58&1.0188&0.8585&0.7696&0.9683\\ 
&SAR-TRN &247.41&0.9407&0.8596&0.7818&0.8911\\
&HTNet               &\color{blue}{258.44}&\color{blue}{1.0116}&\color{blue}{0.8610}&0.7912&0.9679\\ 
&SAR-FAH &\color{red}{1213.76}&\color{red}{0.9905}&\color{red}{0.8611}&\color{red}{0.7970}&\color{red}{0.9896}\\ 
\hline
\multirow{8}{*}{Capella-L4}
&AGSDNet  &409.33&0.9819&0.9943&0.9923&1.2849\\ 
\cline{2-7}
&POTDF &484.10 & 0.9008 &0.9946 & 0.9912 & 0.8977\\
&MuLoG &420.87 & \color{red}{0.9983} & 0.9941 &0.9917 & \color{blue}{0.9925}\\
&SAR-CNN &2228.79&1.0103&0.9943&0.9921&0.9877\\
&SAR-CAM &598.46&1.0068&0.9941&0.9913&0.9912\\ 
&SAR-TRN &1316.61&0.9888&\color{blue}{0.9956}&\color{red}{0.9926}&0.9885\\
&HTNet               &\color{red}{3142.57}&1.0056&0.9951&\color{blue}{0.9923}&0.9864\\ 
&SAR-FAH &\color{blue}{2517.38}&\color{blue}{1.0046}&\color{red}{0.9961}&0.9921&\color{red}{0.9964}\\ 
\hline
\multirow{8}{*}{Iceye-L10}
&AGSDNet & 11099.97&0.9920&0.9686&0.9686&1.1064\\
&POTDF &7709.54 & 0.9483 & 0.9633 & 0.9668 & 0.9786\\
&MuLoG &5137.75 & 0.9959 & \color{blue}{0.9747} & \color{red}{0.9732} & 1.0049\\
&SAR-CNN &3880.31&0.9811&0.9692&0.9676&\color{red}{0.9971}\\
&SAR-CAM &18013.44&\color{red}{0.9965}&0.9670&0.9673&0.9920\\ 
&SAR-TRN &24775.88&0.9303&0.9656&0.9670&0.9366\\
&HTNet               &\color{red}{30466.81}&0.9900&0.9667&0.9674&0.9869\\ 
&SAR-FAH &\color{blue}{25681.02}&\color{blue}{1.0091}&\color{red}{0.9760}&\color{blue}{0.9697}&\color{blue}{1.0034}\\ 
\hline
\end{tabular}
\end{table}
To evaluate the practical value of SAR despeckling methods, we conducted experiments on real SAR images with varying numbers of looks. In the absence of clean references, performance was assessed using both visual performance and no-reference metrics. As shown in figures \ref{fig:real L1}-\ref{fig:real L10}, the proposed SAR-FAH model consistently achieves superior visual results across all real SAR image. It effectively removes speckle noise while preserving critical structural details such as edges, corners, and textures, significantly outperforming comparison methods, which often suffer from over-smoothing, residual noise, or loss of detail.\par
For quantitative evaluation, 4 no-reference QIs are used to evaluate the quality of despeckle images, including equivalent number of looks (ENL) \cite{xieSARSpeckleReduction2002}, mean of image (MoI) \cite{dimartinoBenchmarkingFrameworkSAR2014}, mean of ratio (MoR) \cite{dimartinoBenchmarkingFrameworkSAR2014}, and edge-preservation degree based on the ratio of average (EPD-ROA) along the horizontal direction (HD) and vertical direction (VD) \cite{fengSARImageDespeckling2011}. To analyse the despeckling performance of homogeneous regions, ENL and MoI are calculated in green box of real-SAR images with size of  $10\times 10$, $20\times 20$, and $20\times 20$, respectively. EPD-ROA is calculated to evaluate the edge preservation in heterogeneous regions, which are boxed in blue with size of $40\times 40$ as shown in figure \ref{fig:realimg}. Besides, MoR is calculated to determine the mean preservation of real-SAR image to evaluate overall performance.\par
Quantitative results further confirm the advantages of SAR-FAH. As summarized in table \ref{tab:REAL}, it achieves the best or second-best scores across multiple no-reference QIs. Notably, SAR-FAH excels in both homogeneous region smoothing and structural preservation, with MoR values closest to 1, indicating excellent radiometric consistency. These results demonstrate that SAR-FAH robustly balances speckle suppression and detail retention across diverse real SAR scenarios, underscoring its strong practical applicability.\par
\subsection{Ablation study}
An ablation study on the UCL dataset analyses the efficiency of main structures of SAR-FAH shown in figure \ref{fig:process}. We test on $21$ test images to evaluate them on four QIs with learned parameters.\par
\begin{table}[hbt]
	\setlength\tabcolsep{1pt}
\centering
\caption{The ablation study of main structures in SAR-FAH.}\label{tab:abastru}
\begin{tabular}{lccccc}
\hline
\multicolumn{1}{l}{Method} & \multicolumn{1}{l}{Parm}$\downarrow$ & \multicolumn{1}{l}{PSNR}$\uparrow$ &\multicolumn{1}{l}{SSIM}$\uparrow$ 
&\multicolumn{1}{l}{MAE}$\downarrow$&\multicolumn{1}{l}{GSSIM}$\uparrow$\\
\hline
w/o DCN in DASS&7.40&\color{blue}{23.51}&\color{blue}{0.6198}&\color{blue}{12.3688}&\color{blue}{0.5510}\\
w/o DASS in LFSP-ODE &6.92&23.25&0.5963&12.8012&0.5370\\
w/o DASS in HFDE &6.67&22.87&0.5743&13.3008&0.5293\\
w/o DASS &\color{blue}{6.08}&23.40&0.6046&12.5704&0.5409\\
shared HFDE&\color{red}{3.71}&23.44&0.6162&12.5218&0.5473\\
SAR-FAH &7.32&\color{red}{23.61}&\color{red}{0.6260}&\color{red}{12.2443}&\color{red}{0.5540} \\
\hline
\end{tabular}
\end{table}
Table \ref{tab:abastru} lists the influence of key modules in the proposed SAR-FAH framework. The DASS module extracts both detailed and global features, which are dynamically fused to enhance structural preservation in both the LFSP-ODE and HFDE modules. When using $1\times 1$ convolution to fuse features instead of DCN, it leads to a slight degradation in speckle reduction performance, along with an increase in the number of parameters. Ablation studies further reveal that removing the DASS module from LFSP-ODE, HFDE, or both results in a decline across all QIs, which indicates that the absence of joint local and global feature extraction leads to excessive smoothing and weakened structural retention. Additionally, in the proposed SAR-FAH, we use same but un-shared HFDE to smooth noise in three high-frequency denoising. Experiment shows that using a shared HFDE module fails to fully remove speckle noise. It implies that although high-frequency components exhibit similar noise distributions, they contain distinct structural characteristics that require specialized, non-shared processing modules for effective denoising.\par
\begin{table}[hbt]
	\setlength\tabcolsep{1pt}
\centering
\caption{The ablation study of NODE framework in LFSP-ODE.}\label{tab:abanode}
\begin{tabular}{lccccc}
\hline
\multicolumn{1}{l}{Method} & \multicolumn{1}{l}{Parm}$\downarrow$ & \multicolumn{1}{l}{PSNR}$\uparrow$ &\multicolumn{1}{l}{SSIM}$\uparrow$ 
&\multicolumn{1}{l}{MAE}$\downarrow$&\multicolumn{1}{l}{GSSIM}$\uparrow$\\
\hline
w/o NODE &\color{red}{7.32}&23.18&0.5966&12.8905&0.5348\\
Residual substitution&\color{blue}{11.51}& \color{blue}{23.53}&\color{blue}{0.6199}&\color{blue}{12.3318} &0.5487\\
SAR-FAH &\color{red}{7.32}&\color{red}{23.61}&\color{red}{0.6260}&\color{red}{12.2443}&\color{red}{0.5540}\\
\hline
\end{tabular}
\end{table}
Table \ref{tab:abanode} summarizes the impact of the NODE framework, which achieves improved performance and enhanced structural fidelity with fewer parameters. Removing the NODE structure from the LFSP-ODE module, while keeping the total parameter count unchanged, leads to significant degradation in structural preservation. Replacing NODE with discrete residual connections of the same depth substantially increases the number of parameters without yielding performance gains. These results demonstrate that the NODE framework in LFSP-ODE effectively contributes to better structural preservation as well as less  computational burden.\par 
\begin{table}[hbt]
	\setlength\tabcolsep{3pt}
\centering
\caption{The ablation study of DeforConv in HFDE.}\label{tab:abah}
\begin{tabular}{ccccccc}
\hline
\multicolumn{1}{c}{Encoder}&\multicolumn{1}{c}{Decoder}& \multicolumn{1}{l}{Parm}$\downarrow$ & \multicolumn{1}{l}{PSNR}$\uparrow$ &\multicolumn{1}{l}{SSIM}$\uparrow$ 
&\multicolumn{1}{l}{MAE}$\downarrow$&\multicolumn{1}{l}{GSSIM}$\uparrow$\\
\hline
\ding{55}&\ding{55} &\color{red}{7.23}&23.09&0.5920&12.9606&0.5358\\
\ding{51}&\ding{51} &7.41&23.48&0.6136&12.4111&\color{blue}{0.5473}\\
\ding{51}&\ding{55}&\color{blue}{7.32}&\color{blue}{23.52}&\color{blue}{0.6143}&\color{blue}{12.3487}&0.5465\\
\ding{55}&\ding{51} &\color{blue}{7.32}&\color{red}{23.61}&\color{red}{0.6260}&\color{red}{12.2443}&\color{red}{0.5540}\\
\hline
\end{tabular}
\end{table}
Table \ref{tab:abah} analyses the role of DeforConvs in the asymmetric HFDE module. It is clear that removing DeforConvs from the HFDE leads to a noticeable decline in denoising performance, characterized by reduced effectiveness in high-frequency regions and excessive smoothing of structural details. The asymmetry arises since DeforConvs are used only in the decoder part of the HFDE. When we also integrate them into the encoder, forming a symmetric architecture, denoising performance degrades further and the ability to preserve high-frequency information is impaired. Similarly, using DeforConvs only in the encoder also yields suboptimal results.\par
Tables \ref{tab:deuhomo} and \ref{tab:deuheme} demonstrate the effectiveness of the specialized frequency-domain processing networks LFSP-ODE and HFDE. For evaluation, we constructed two separate datasets: one with limited textures and another with rich textures. Each dataset contains 450 images from 15 categories, and experiments were conducted on 15 representative test images. To accelerate comparison, the feature channel dimension $C$ was set to 64. As shown in table \ref{tab:deuhomo}, applying the HFDE to low-frequency denoising causes significant loss of image structure. Conversely, replacing the HFDE with LFSP-ODE for high-frequency denoising results in over-smoothed. 
\begin{table}[hbt]
	\setlength\tabcolsep{3pt}
\centering
\caption{The ablation study of LFSP-ODE and HFDE for low-frequency bands despeckling.}\label{tab:deuhomo}
\begin{tabular}{lccccc}
\hline
\multicolumn{1}{l}{Method}&\multicolumn{1}{l}{Parm}$\downarrow$ & \multicolumn{1}{l}{PSNR}$\uparrow$ &\multicolumn{1}{l}{SSIM}$\uparrow$ 
&\multicolumn{1}{l}{MAE}$\downarrow$&\multicolumn{1}{l}{GSSIM}$\uparrow$\\
\hline
HFDE + HFDE &\color{red}{1.57}&24.41&0.7078&11.1764&0.7807\\
LFSP-ODE + HFDE &1.90&\color{red}{26.22}&\color{red}{0.7491}&\color{red}{9.1750}&\color{red}{0.8048}\\
\hline
\end{tabular}
\end{table}
\begin{table}[hbt]
	\setlength\tabcolsep{3pt}
\centering
\caption{The ablation study of LFSP-ODE and HFDE for high-frequency bands despeckling.}\label{tab:deuheme}
\begin{tabular}{lccccc}
\hline
\multicolumn{1}{l}{Method} &\multicolumn{1}{l}{Parm}$\downarrow$& \multicolumn{1}{l}{PSNR}$\uparrow$ &\multicolumn{1}{l}{SSIM}$\uparrow$ 
&\multicolumn{1}{l}{MAE}$\downarrow$&\multicolumn{1}{l}{GSSIM}$\uparrow$\\
\hline
LFSP-ODE + LFSP-ODE &2.01&20.63&0.4790&18.4267&0.4502\\
LFSP-ODE + HFDE &\color{red}{1.90}&\color{red}{20.66}&\color{red}{0.4914}&\color{red}{18.2780}&\color{red}{0.4586}\\
\hline
\end{tabular}
\end{table}
\subsection{Model analysis}
\subsubsection{Parameter analysis}
The NODE framework is used in LFSP-ODE module to remove noise in low-frequency. In this framework, the hyper-parameter $N$ in equation (\ref{eq:node-low}) is important to control model complexity and impact despeckling performance. As tabulated in table \ref{tab:aba n}, we systematically evaluate its impact on UCL dataset. It is seen that as $N$ decreases, the testing time increases and the performance comes better firstly and then decreases. The best choice is $N = 4$ to balance efficiency and performance.\par 
\begin{table}[hbt]
	\setlength\tabcolsep{3pt}
\centering
\caption{The parameter analysis of $N$ in LFSP-ODE.}\label{tab:aba n}
\begin{tabular}{llccccc}
\hline
\multicolumn{1}{l}{Parm}$\downarrow$&\multicolumn{1}{l}{N}&  \multicolumn{1}{l}{PSNR}$\uparrow$ &\multicolumn{1}{l}{SSIM}$\uparrow$ 
&\multicolumn{1}{l}{MAE}$\downarrow$ &\multicolumn{1}{l}{GSSIM}$\uparrow$ &\multicolumn{1}{l}{Time(s)}$\downarrow$\\
\hline
\multirow{6}{*}{7.32}
&1 &23.42&0.6152&12.4994&0.5510&\color{red}{1.41}\\
&2 &23.55&0.6179&12.3250&0.5498&\color{blue}{1.95}\\
&3 &23.58&\color{blue}{0.6222}&12.3159&0.5513&2.64\\
&4 &\color{red}{23.61}&\color{red}{0.6260}&\color{red}{12.2443}&\color{red}{0.5540}&3.10\\
&5 &\color{blue}{23.60}&0.6230&\color{blue}{12.2531}&0.5505&3.60\\
&6 &23.59&\color{blue}{0.6237}&12.2684&\color{blue}{0.5532}&4.27\\
\hline
	\end{tabular}
\end{table}
\subsection{Computational efficiency} 
\begin{figure}
\centering
\setlength{\abovecaptionskip}{0cm}
\includegraphics[width=0.4\textwidth]{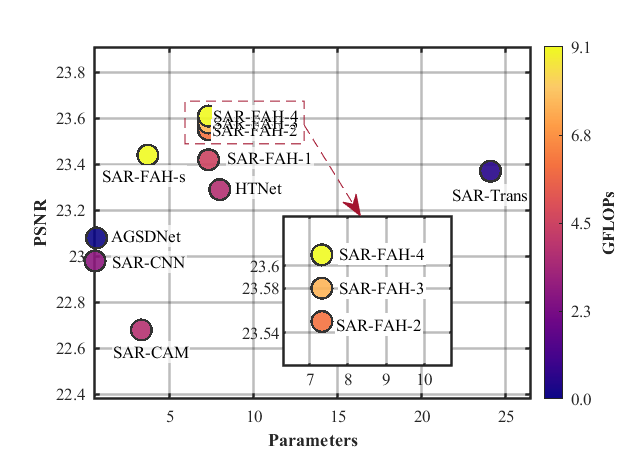}
\caption{Efficiency on comparative deep-learning methods, including PSNR, parameters, and GFLOPs, on UCL dataset.}
\label{fig:efficiency}
\end{figure}

Balancing computational efficiency with despeckling performance is essential for the practical applicability of SAR despeckling models. As illustrated in figure \ref{fig:efficiency}, we compare the denoising effectiveness, parameter, and GFLOPs of learning-based models. It includes the proposed SAR-FAH-s (with shared HFDE modules), variants SAR-FAH-$i$ (where $N=i$ in the LFSP-ODE module), and other comparative approaches. Our model also achieves competitive results with fewer parameters while sharing HFDE parameters. Furthermore, by adjusting the value of $N$, we can flexibly control computational complexity. Notably, even when $N=1$, the model maintains strong despeckling performance. These results confirm that the proposed SAR-FAH achieves a favorable balance between despeckling efficacy and computational cost, establishing its strong practical value in despeckling tasks.\par
\section{Conclusion}
\label{sec:conclusion}
This paper presents a novel a Frequency-Adaptive Hybrid model (SAR-FAH) based on NODEs for SAR despeckling, which achieves a better balance between noise suppression and structure preservation. SAR-FAH employs wavelet transform to decouple the image into different frequency sub-bands and theoretically analyze the statistical distribution of different sub-bands, thereby enabling frequency-adaptive processing. For low-frequency homogeneous regions, an LFSP-ODE module based on the NODE framework is introduced to achieve effective noise suppression in uniform areas while preserving structural integrity without introducing artifacts. Furthermore, by adjusting the NODE framework, the computational efficiency of the model can be flexibly controlled to meet different requirements. For high-frequency sub-bands, an asymmetric U-shaped module derived from U-Net captures multi-scale texture information and incorporates DeforConv to extract edges and textures of diverse shapes, thereby maintaining sharp structures and fine details. Furthermore, a DASS module is designed to integrate both attention mechanisms and VMamba structures, dynamically fusing local and global features to enhance overall performance and structural preservation. Comprehensive evaluations on synthetic SAR, texture, and real SAR datasets, demonstrates superior despeckling performance, effectively removing noise while preserving edges and textures well.
\bibliographystyle{IEEEtran}
\bibliography{citeb}

\newpage

%


%


\end{document}